\documentclass[11pt]{article}
\usepackage[preprint]{acl}

\usepackage{times}
\usepackage{latexsym}
\usepackage[T1]{fontenc}
\usepackage[utf8]{inputenc}
\usepackage{microtype}
\usepackage{inconsolata}
\usepackage{graphicx}
\usepackage{booktabs}
\usepackage{amsfonts}
\usepackage{enumitem}
\usepackage{amsmath}
\usepackage{float}
\usepackage{tcolorbox}
\usepackage{listings}
\usepackage{xcolor}

\lstdefinelanguage{json}{
    basicstyle=\ttfamily\small,
    showstringspaces=false,
    breaklines=true,
    morestring=[b]",
    stringstyle=\color{blue!70!black},
    commentstyle=\color{gray},
}

\title{TopoAgent: A Structure-Aware Perception-to-Reasoning Framework for Diagram-to-Graph Topology Extraction with Large Vision-Language Models}

\author{
Bangwei Guo$^{1}$ \quad
Xujiang Zhao$^{2}$ \quad
Yanchi Liu$^{3}$ \quad
Wei Cheng$^{3}$ \\[2pt]
\textbf{Shengyu Chen}$^{3}$ \quad
\textbf{Dongyue Li}$^{4}$ \quad
\textbf{Masaharu Morimoto}$^{5}$ \\[2pt]
\textbf{Takayuki Kuroda}$^{5}$ \quad
\textbf{Dimitris Metaxas}$^{1}$ \quad
\textbf{Haifeng Chen}$^{3}$ \\[4pt]
$^{1}$Rutgers University \quad
$^{2}$Meta \quad
$^{3}$NEC Labs America \\
$^{4}$University of Electro-Communications \quad 
$^{5}$NEC Corporation \\
\texttt{bg654@rutgers.edu, zxj32@meta.com} 
}

\begin{document}
\maketitle

\begin{abstract}
Diagram-to-graph topology extraction aims to extract a graph of entities and their connections from a structural diagram. This task remains challenging for current vision-language models because it requires both fine-grained perceptual grounding and topology-aware reasoning with global consistency. We present \textbf{TopoBench-180}, a human-verified benchmark for diagram-to-graph topology extraction, and \textbf{TopoAgent}, a structure-aware perception-to-reasoning framework for reliable topology extraction using large vision-language models. TopoBench-180 contains 180 structural diagrams spanning Web-style and Network-style categories, paired with canonical graph annotations. TopoAgent progressively extracts the target graph by combining grounded perception, global structural priors, canonical node inventory construction, node-centric local-to-global relation reasoning, and topological consistency enforcement. Experiments on TopoBench-180 show that TopoAgent outperforms strong vision-language model baselines and recent visual reasoning frameworks, especially on edge extraction. More broadly, this work fills an important gap in multimodal structured understanding by establishing a benchmark and framework for diagram-to-graph topology extraction. The benchmark and associated resources will be publicly released at \url{https://huggingface.co/datasets/WayneGuo0011/TopoBench-180}.
\end{abstract}

\section{Introduction}

\begin{figure}[t]
\centering
\includegraphics[width=\columnwidth]{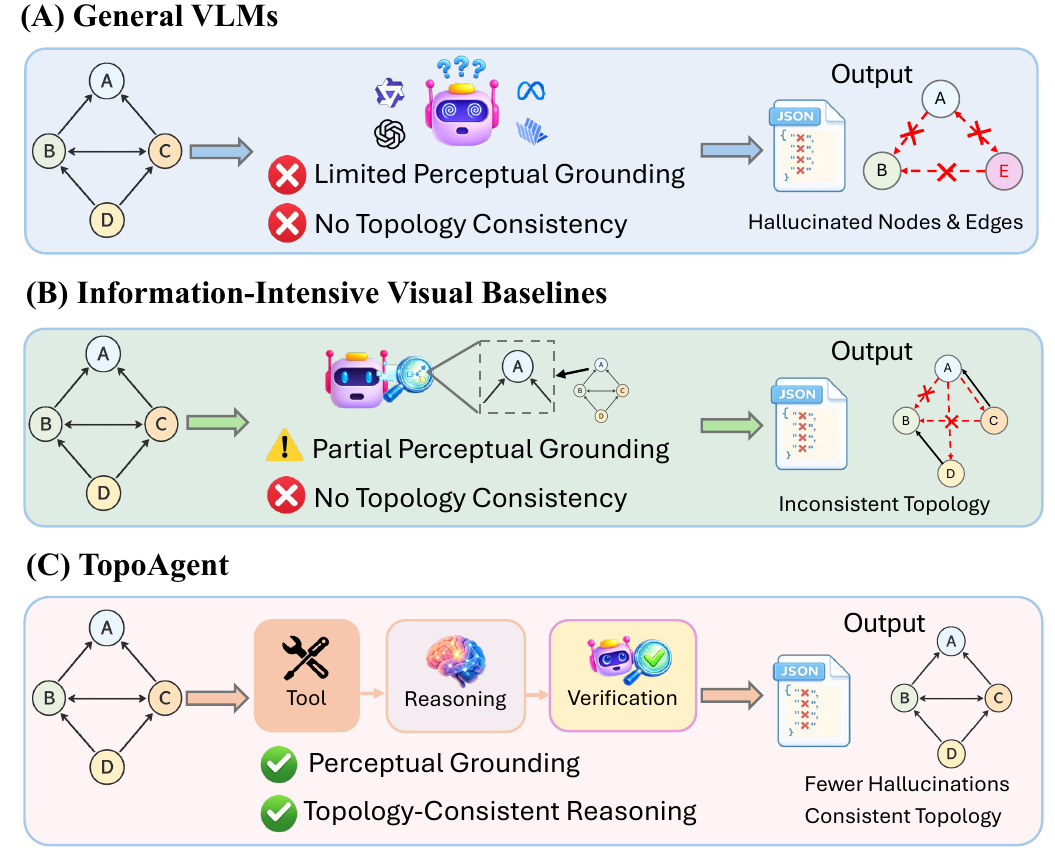}
\caption{Motivating comparison for diagram-to-graph topology extraction. General-purpose VLMs often miss fine-grained visual relations or produce topologically inconsistent graphs, while prior visual reasoning baselines only partially address these challenges. In contrast, TopoAgent combines grounded perception with topology-aware reasoning to improve both local relation extraction and global graph consistency.}
\label{fig:reasoning_example}
\end{figure}

Technical and scientific diagrams convey rich relational structure through labeled entities, visual connectors, and geometric layout. Extracting their topology is important for downstream applications including graph-based reasoning~\cite{besta2024graph}, knowledge graph construction~\cite{zhu2024llms}, and scientific understanding~\cite{lu2022learn}. However, despite recent progress in large vision-language models (VLMs) on multimodal tasks such as chart understanding~\cite{masry2022chartqa, liu2025small}, visual question answering~\cite{yue2024mmmu, li2025survey}, and structured information extraction~\cite{li2025chain, 11298739}, diagram-to-graph topology extraction remains underexplored. Unlike prior work that mainly answers specific queries or produces coarse textual descriptions~\cite{ahn2024large}, this task requires extracting the topology of a diagram as a structured graph. It is substantially more demanding because the extracted topology must be globally structured and topologically consistent.

% Recent advances in large vision-language models (VLMs) have demonstrated impressive performance on multimodal tasks such as chart understanding~\cite{masry2022chartqa, liu2025small}, visual question answering~\cite{yue2024mmmu, li2025survey}, and structured information extraction~\cite{li2025chain, 11298739}. While prior work has primarily focused on question answering or caption-level understanding~\cite{ahn2024large}, relatively limited research has addressed diagram-to-graph topology extraction, a substantially more challenging task that requires reconstructing the complete relational structure of a diagram rather than answering a single query. Despite its clear importance for downstream applications such as graph-structured reasoning~\cite{besta2024graph}, knowledge graph construction~\cite{zhu2024llms}, and scientific reasoning~\cite{lu2022learn}, this problem remains underexplored, with few systematic benchmarks or robust extraction methods available.

The difficulty of this task arises from two main technical challenges. First, it requires precise perceptual grounding in information-dense environments, where textual labels, geometric primitives, and thin connectors are tightly interleaved across multi-scale layouts~\cite{masry2022chartqa, yue2024mmmu}. Even subtle ambiguities in connector attachment or directionality can lead to serious topology errors. Second, the task requires globally consistent reasoning rather than isolated local predictions. Models must determine which entities are connected while maintaining consistency across node--edge assignments; otherwise, local errors can propagate into missing edges, hallucinated nodes, and inconsistent graphs~\cite{huang2024visual}. Progress on this problem has also been limited by the lack of dedicated benchmarks. Constructing high-quality diagram--graph pairs requires exhaustive node identification, direction-aware edge labeling, and careful human verification, which are often absent from existing large-scale datasets.

Existing approaches to information-intensive visual reasoning only partially address the demands of diagram-to-graph topology extraction. Region-based or ``zoom-in'' methods improve fine-grained perception through localized decomposition~\cite{zheng2025deepeyes,li2025chain,wang2025pixel}, but do not explicitly enforce topology-consistent extraction. Multi-step reasoning frameworks improve sequential inference in general multimodal tasks~\cite{lu2022learn,Xu_2025_ICCV,wang2025multimodal}, but are not tailored to the geometric and structural demands of topological diagrams, often missing subtle yet critical cues such as connector crossings and long-range dependencies. Moreover, most established visual reasoning benchmarks~\cite{yue2024mmmu, masry2022chartqa, liu2024mmbench} focus on question answering or caption generation rather than full diagram-to-graph extraction. These limitations motivate a structure-aware extraction framework and a dedicated benchmark, as summarized in Figure~\ref{fig:reasoning_example}.

To address these limitations, we introduce \textbf{TopoBench-180}, a human-verified benchmark specifically designed for diagram-to-graph topology extraction, together with \textbf{TopoAgent}, a structure-aware perception-to-reasoning framework for reliable topology extraction with large VLMs. TopoBench-180 provides high-quality diagram--graph pairs with canonical annotations and diverse topology patterns, including directionality and densely connected relational structures, enabling rigorous evaluation of topology extraction under structurally challenging settings. TopoAgent begins with a low-level perceptual grounding stage, which decomposes the input diagram into candidate regions, associates them with OCR-recognized text, and extracts region-level primitive summaries for downstream reasoning. It then performs four coordinated reasoning stages: (i) global attribute inference, (ii) canonical node grounding, (iii) node-centric relation reasoning, and (iv) topological consistency enforcement. This design progressively constructs the final graph from local evidence to global consistency. Experiments on TopoBench-180 show that TopoAgent consistently outperforms strong closed-source VLMs, open-source VLMs, and recent visual reasoning frameworks.

% To address these limitations, we introduce a structure-aware framework for diagram-to-graph topology extraction, together with a new human-verified benchmark. We first construct \textbf{TopGraph-180}, a dedicated evaluation dataset comprising high-quality diagram–graph pairs with canonical structured annotations. The benchmark emphasizes explicit relational topology, including directionality, multi-entity interactions, and densely connected graph structures, thereby enabling systematic evaluation of topology reconstruction quality.

% Building upon this benchmark, we propose \textbf{TopoAgent}, a multi-stage reasoning framework inspired by the practical workflow of human topology annotation. In manual annotation, annotators typically first identify all nodes and their textual labels, then examine each node to determine its connected edges and relation types, and finally verify the overall structural consistency of the constructed graph. TopoAgent mirrors this staged process by explicitly disentangling perceptual grounding from topology reasoning and structural validation. Rather than generating the entire graph in a single end-to-end pass, the framework decomposes the task into structured sub-processes: low-level visual grounding for node identification, topology-aware relational reasoning for edge construction, and schema-constrained validation for global consistency checking. This decomposition enables more controlled and interpretable graph reconstruction.

Our contributions are summarized as follows:
\begin{itemize}[leftmargin=*, itemsep=1pt, topsep=1pt]
    \item To the best of our knowledge, this is the first work to systematically study diagram-to-graph topology extraction in the context of large vision-language models.
    \item We introduce TopoBench-180, a dedicated human-verified benchmark with canonical graph annotations for rigorous evaluation of topology extraction.
    \item We propose TopoAgent, a structure-aware perception-to-reasoning framework that combines grounded region summarization, staged topology reasoning, and topological consistency enforcement for topology extraction.
    \item Experiments on TopoBench-180 show that TopoAgent consistently outperforms strong VLM baselines and recent visual reasoning frameworks.
\end{itemize}

Together, our work advances topology-aware diagram understanding by pairing a dedicated benchmark with a structure-aware framework for reliable topology extraction.

\section{Related Work}

\paragraph{Benchmarks for Diagram and Structured Visual Understanding.}
A growing number of benchmarks have been proposed for reasoning over structured or information-dense visual inputs. 
AI2D~\cite{kembhavi2016diagram} and AI2D-RST~\cite{hiippala2021ai2d} provide annotated science diagrams for question answering and discourse-level structural analysis. 
Other datasets focus on visual reasoning over structured graphics such as charts, plots, and infographics, including ChartQA~\cite{masry2022chartqa}, PlotQA~\cite{Methani_2020_WACV}, and InfographicVQA~\cite{mathew2022infographicvqa}. 
More recent multimodal benchmarks, such as MMBench~\cite{liu2024mmbench}, ChartMuseum~\cite{tang2025chartmuseum}, and MMMU~\cite{yue2024mmmu}, further evaluate the reasoning abilities of vision-language models across diverse visual tasks. 
However, these benchmarks primarily focus on question answering, caption generation, or general multimodal reasoning, rather than full diagram-to-graph topology extraction with canonical graph annotations.

\paragraph{Enhancing Spatial and Geometric Understanding in VLMs.}
Recent work has sought to improve the spatial perception of vision-language models for structured visual inputs. 
Some approaches modify model architectures to incorporate explicit spatial cues such as bounding boxes~\cite{liu2024chain}, segmentation masks~\cite{guo2024regiongpt}, or OCR-extracted text regions~\cite{li2025chain}. 
Other methods improve spatial reasoning through additional supervision or instruction tuning, especially for charts, documents, and diagrams~\cite{zhang2024tinychart, meng2024chartassistant, fan2024read, zheng2025deepeyes}. 
While these methods improve fine-grained visual grounding, they are not specifically designed for diagram-to-graph topology extraction, where local perceptual accuracy must be coupled with globally consistent structural reasoning.

\paragraph{Vision--Language Model Reasoning for Structural Visual Understanding.}
Recent research has also explored improving VLM reasoning through structured reasoning paradigms. 
Prompt-based approaches encourage models to generate intermediate reasoning steps through chain-of-thought prompting~\cite{xu2025llava,shao2024visual}, while reinforcement learning methods further optimize these reasoning trajectories~\cite{huang2025vision,ma2025deepperception,guo2025deepseek}. 
More recently, agentic frameworks treat VLMs as planners that decompose tasks and invoke visual tools such as zooming or region selection during reasoning~\cite{wu2024v, hu2024visual,yang2025machine}. 
However, these methods are not designed for topology extraction from diagrams, where precise entity grounding, consistent node--edge assignment, and graph-level structural constraints must be handled jointly.

\begin{figure}[ht!]
\centering
\includegraphics[width=\columnwidth]{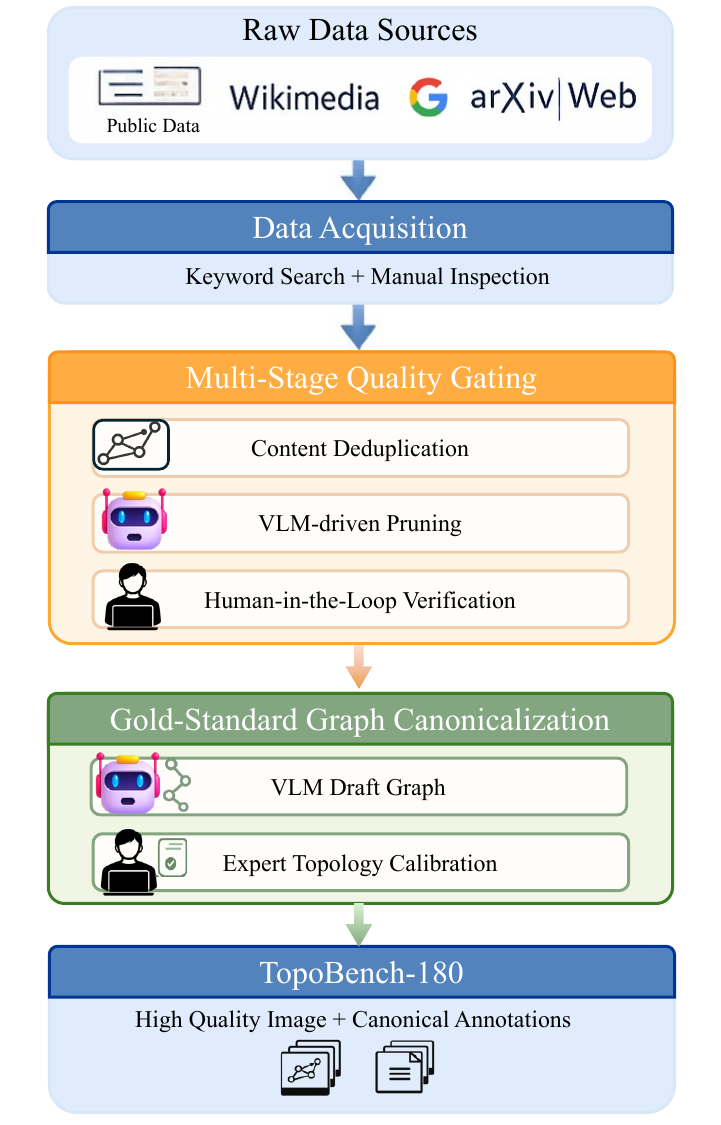}
\caption{
Pipeline for constructing the \textbf{TopoBench-180} benchmark. Candidate diagrams are collected from existing public datasets, public web sources, and curated internal sources. The initial pool of raw images is then filtered through a multi-stage quality gating pipeline consisting of content-based de-duplication, VLM-driven pruning, and annotator cross-audit. For each retained diagram, a draft graph is first generated under a unified topology schema and then manually verified and refined by annotators to produce the final canonical graph annotation.
}
\label{fig:dataset_pipeline}
\end{figure}

\section{TopoBench-180}
\label{sec:dataset}
\paragraph{Data Sources.}
We introduce and publicly release TopoBench-180\footnote{
\url{https://huggingface.co/datasets/WayneGuo0011/TopoBench-180}}, a human-verified benchmark for diagram-to-graph topology extraction (Figure~\ref{fig:dataset_pipeline}). It is constructed from a combination of existing public datasets, curated web collections, and internal diagram collection. We use diagrams from AI2D~\cite{kembhavi2016diagram} and the Network Topology Diagram Computer Vision Dataset\footnote{\url{https://universe.roboflow.com/project-v2jjh/network-topology-diagram}} as primary sources, and further expand structural diversity with samples gathered from publicly accessible resources, including Google image search, Wikipedia/Wikimedia Commons, arXiv, and related repositories.

To ensure broad structural coverage, we collect diagrams using a diverse set of topology-oriented keywords and source-specific queries, followed by manual review to retain samples with diverse structural patterns. This process allows us to include both relational diagrams (e.g., food webs and biological interaction graphs) and network or system diagrams with complex connectivity patterns. In total, the initial pool contains over 6,000 raw images, including 3,088 from AI2D, 1,627 from Wikipedia- and arXiv-based sources, 1,605 from Google Image Search, and additional internal collections. For benchmark samples whose licenses permit redistribution, the released dataset directly includes the corresponding images. For the remaining samples, we provide canonical graph annotations, provenance metadata, source links, and preprocessing scripts to support reproducibility.

\subsection{Multi-Stage Quality Gating}
Raw web-scale collections contain redundancy, visually noisy figures, and structurally trivial cases. To ensure benchmark quality, we adopt a coarse-to-fine quality gating pipeline based on semantic validity, structural quality, annotation feasibility, and diversity. We first perform content-based de-duplication, reducing the initial pool of over 6,000 raw diagrams to 4318 unique candidates. The remaining samples then undergo a VLM-driven pruning with GPT-4o, where the VLM serves as a coarse filter for explicit multi-entity topology, relational clarity, and suitability for the annotation. This stage further reduces the pool to 741 structurally valid candidates. Finally, the retained samples are cross-audited by two annotators under a predefined annotation rubric to verify semantic validity, topological completeness, annotation feasibility, and structural diversity. This final curation stage yields 180 samples for canonical graph annotation. Detailed filtering prompts, criteria, and stage-wise statistics are provided in the Appendix.

% For each retained diagram, we construct a canonical graph annotation through a coarse-to-fine annotation pipeline. We first generate a draft structured graph using GPT-4o under a unified topology extraction schema, producing an initial adjacency-level representation of nodes and relations. This draft annotation is then manually refined by expert annotators through topology calibration, including correction of missing or spurious edges, edge directionality, node identity consistency, and node type assignments. All finalized annotations are serialized in structured \texttt{JSON} format under a canonical schema, and all images are standardized and stored in \texttt{PNG} format. 
% The resulting \textbf{TopoBench-180} benchmark contains 180 high-quality diagram--graph pairs, including 54 web-style diagrams and 126 network or complex system diagrams. 
\subsection{Gold-Standard Graph Canonicalization}
For each retained diagram, we construct a canonical graph annotation through a coarse-to-fine annotation pipeline. We first use GPT-4o to generate a draft structured \texttt{JSON} under a unified topology extraction schema, serving only as an initial annotation scaffold. Annotators then manually verify and refine all node and relation entries through topology calibration, including correction of missing or spurious edges, edge directionality, and node identity consistency. The final annotation is therefore not taken directly from model output, but established through human-guided canonicalization under a unified graph schema. All finalized annotations are serialized in canonical \texttt{JSON} format, and all images are standardized and stored in \texttt{PNG} format. The resulting TopoBench-180 benchmark contains 180 high-quality diagram--graph pairs, including 54 Web-style diagrams and 126 Network-style diagrams, supporting evaluation across distinct diagram styles and link directionality patterns.

\begin{figure*}[t]
    \centering
    \includegraphics[width=\textwidth]{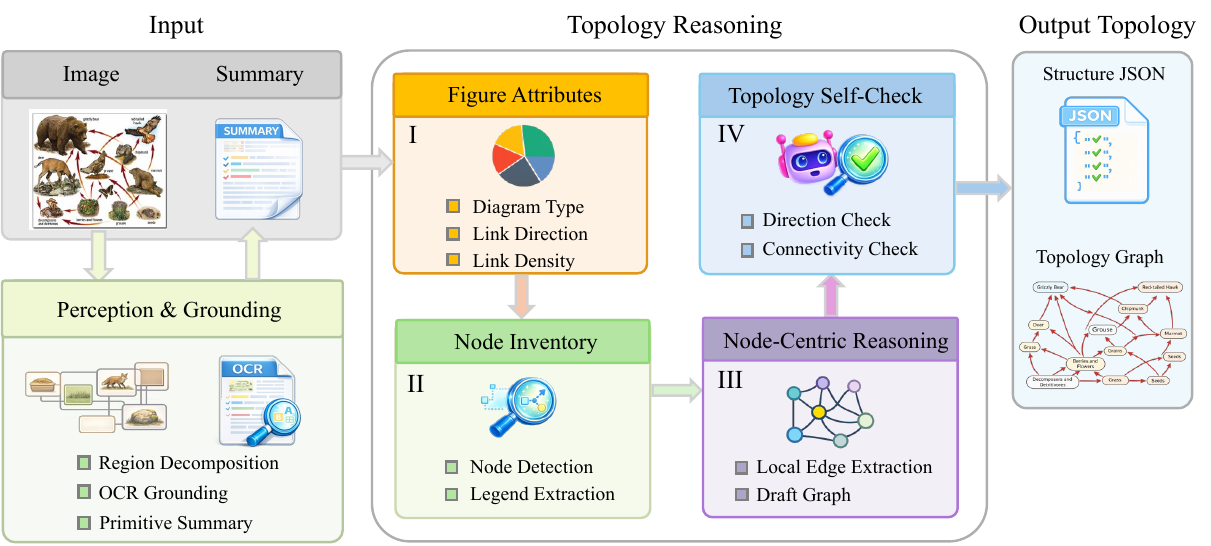}
    \caption{
    Overview of the proposed TopoAgent framework for diagram-to-graph topology extraction.
    The pipeline begins with a perception and grounding stage that decomposes the input diagram into grounded regions, aligns them with OCR-recognized text, and constructs region-level structured summaries. These grounded representations are then passed to a local-to-global topology reasoning pipeline, which produces a structured topology JSON representing the extracted graph.
    }
    \label{fig:topoagent_pipeline}
\end{figure*}

\section{Method}

\subsection{Problem Formulation}
Given a structural diagram image $I$, our goal is to extract a canonical topology graph
\begin{equation}
G = (V, E),
\end{equation}
where $V$ denotes the set of diagram nodes and $E$ denotes the set of edges between them. Each node $v \in V$ corresponds to a canonical entity grounded in the diagram. Each edge is represented in canonical directed form as
\begin{equation}
e = (\texttt{source}, \texttt{target}),
\end{equation}
where \texttt{source} and \texttt{target} denote the two endpoint nodes of a connection. For bidirectional or undirected connections, both directed edges $(u, v)$ and $(v, u)$ are included in the canonical graph representation. Auxiliary cues such as relation semantics and global structural attributes may be used during reasoning, while evaluation is performed on the canonical node and edge structure.

\subsection{Overview of the TopoAgent Framework}

TopoAgent formulates diagram-to-graph topology extraction through a perception-to-reasoning pipeline that progressively constructs grounded local evidence to a globally coherent graph. As illustrated in Figure~\ref{fig:topoagent_pipeline}, the framework consists of five coordinated stages:
\begin{enumerate}
    \item \textbf{Perception \& Grounding}: extracts grounded visual regions, OCR-aligned text, and region-level primitive summaries.
    \item \textbf{Global Figure Attribute Inference}: predicts high-level structural priors for downstream reasoning.
    \item \textbf{Canonical Node Inventory Construction}: builds a stable symbolic node inventory and legend summary.
    \item \textbf{Node-Centric Local-to-Global Relation Reasoning}: performs node-anchored relation synthesis under shared symbolic context.
    \item \textbf{Topological Consistency Enforcement}: consolidates draft relations into a valid graph under explicit structural constraints.
\end{enumerate}

\subsection{Perception \& Grounding}

TopoAgent begins with a perception and grounding stage that converts the input diagram from raw pixels into grounded structured evidence for downstream topology reasoning. This stage decomposes the diagram into candidate visual regions, aligns them with textual content, and constructs region-level structured summaries for later symbolic reasoning.

\paragraph{Region Decomposition and OCR Grounding.}
Given an input diagram image $I \in \mathbb{R}^{H \times W \times 3}$, we decompose the image into candidate visual regions corresponding to diagram primitives (e.g., nodes, lines, arrows, connectors, and enclosures). Following prior work~\cite{li2025chain}, this process consists of three stages: initialization, splitting, and merging. We first detect connected visual components to obtain coarse candidate regions, then split composite regions using geometric cues (e.g., circles, rectangles, and line segments), and finally merge related fragments based on spatial proximity and structural evidence. We apply PaddleOCR~\cite{cui2025paddleocr30technicalreport,cui2025paddleocrvlboostingmultilingualdocument} to detect text regions and extract their bounding boxes, which provide spatial anchors for region merging. The final output is a set of grounded region--text pairs $\mathcal{R} = \{(r_i, t_i)\}_{i=1}^{N}$, where $r_i$ denotes the $i$-th visual region and $t_i$ its associated text label.

\paragraph{Region-Level Primitive Summarization.}
Given the grounded region--text pairs $\mathcal{R} = \{(r_i, t_i)\}_{i=1}^{N}$, we use a large vision-language model (GPT-4o) to generate a structured summary for each region. For each region $r_i$, the model predicts candidate node labels, geometric attributes (e.g., position, size, and shape), and potential connection primitives such as arrows or line segments, together with their geometric endpoints. These outputs are aggregated into a region-level representation $\mathcal{S} = \{s_i\}_{i=1}^{N}$, where $s_i$ denotes the structured summary of region $r_i$, including its associated text and geometric attributes. The resulting summaries $\mathcal{S}$ provide grounded perceptual evidence for subsequent topology reasoning. Implementation details are provided in the Appendix.

\subsection{Local-to-Global Topology Reasoning}

Given the grounded summaries $\mathcal{S}$ from the perception stage and the input image $I$, TopoAgent performs topology reasoning by progressively constructing a globally coherent graph. The reasoning process proceeds through global figure attribute inference, canonical node inventory construction, node-centric relation reasoning, and final topological consistency enforcement.

\paragraph{Stage I: Global Figure Attribute Inference.}
The first stage predicts a set of global figure attributes
\begin{equation}
A = \{a_{\text{sem}}, a_{\text{dir}}, a_{\text{dens}}\},
\end{equation}
which characterize the diagram semantics, link directionality, and link density, respectively. Concretely, the model classifies the figure as \texttt{web}, \texttt{network}, or \texttt{structural\_diagram}; determines whether links are \texttt{one\_direction} or \texttt{bi\_direction}; and estimates whether inter-node connectivity in the diagram is predominantly \texttt{single\_link} or \texttt{multi\_link}. These attributes serve as high-level structural priors that guide downstream reasoning and reduce ambiguity in relation synthesis.

\paragraph{Stage II: Canonical Node Inventory Construction.}
The second stage constructs a canonical node inventory
\begin{equation}
V = \{v_i\}_{i=1}^{N},
\end{equation}
by consolidating visual evidence from the input image $I$, the OCR-grounded region-level summaries $\mathcal{S}$ into a stable symbolic node list. Each node represents a semantically valid diagram entity grounded in textual labels and supporting visual evidence. In addition to the node set, this stage identifies discarded text regions that do not correspond to valid entities and summarizes legend information when present. This stage establishes a stable symbolic scaffold that separates entity grounding from downstream topology reasoning.

\paragraph{Stage III: Node-Centric Local-to-Global Relation Reasoning.}
A core design of TopoAgent is to avoid predicting all relations jointly under a single global visual context. In dense diagrams, such one-shot reasoning can miss fine-grained connectors such as small arrows, thin lines, or crowded edge attachments. To address this, we adopt a node-centric local-to-global decomposition that reduces full-graph extraction to $N$ node-anchored subproblems. For each focus node $v_i \in V$, the relation reasoning stage takes the image $I$, the grounded summaries $\mathcal{S}$, and the global attributes $A$, and predicts only the relations originating from $v_i$. The resulting node-local edge set is denoted by
\begin{equation}
E_i = \mathcal{F}_{\text{rel}}(v_i, I, \mathcal{S}, A).
\end{equation}
Aggregating all node-local predictions yields a draft edge set
\begin{equation}
E_{\text{draft}} = \bigcup_{i=1}^{|V|} E_i .
\end{equation}

This node-centric decomposition preserves global structural context while enforcing localized visual focus, improving extraction of subtle connections in complex diagrams.

\paragraph{Stage IV: Topological Consistency Enforcement.}
The final stage performs topological consistency enforcement on the draft graph without revisiting the raw image, thereby avoiding additional visual hallucinations during verification. Starting from $E_{\text{draft}}$, it refines the draft graph under explicit structural constraints.

First, it enforces directionality consistency. If the predicted figure attributes indicate bidirectional or undirected connectivity, a valid connection between $v_i$ and $v_j$ must be represented by both $(v_i, v_j)$ and $(v_j, v_i)$ in the canonical graph. If the figure attributes indicate one-directional connectivity, unsupported reciprocal edges are treated as directionality conflicts and resolved conservatively based on the predicted attributes and draft graph structure. Second, it enforces node validity and connectivity consistency. All edge endpoints must belong to the canonical node set $V$, and the final graph is constrained to form a single connected component. Edges or fragments that violate these constraints are removed. The final topology graph is obtained as
\begin{equation}
G^* = \operatorname{SelfCheck}(V, E_{\text{draft}}, A).
\end{equation}

\begin{table*}[t]
\centering
\small
\setlength{\tabcolsep}{6pt}
\begin{tabular*}{\textwidth}{@{\extracolsep{\fill}}l c cccc|cccc}
\toprule
 &  & \multicolumn{4}{c|}{Web-style} 
 & \multicolumn{4}{c}{Network-style} \\
\cmidrule(lr){3-6} \cmidrule(lr){7-10}
Model 
& Params 
& Node F1 & Edge P & Edge R & Edge F1
& Node F1 & Edge P & Edge R & Edge F1 \\
\midrule

\multicolumn{10}{c}{\textit{Closed-source VLMs}} \\
\midrule
GPT-4o 
& -- 
& 93.3 & 37.8  & 30.0  & 32.8
& 85.2 & 57.8 & 47.1 & 53.5  \\
GPT-4o-mini
& -- 
& 90.5 & 34.0 & 31.6 & 31.6
& 72.0 & 46.6 & 34.6 & 38.2 \\

\midrule
\multicolumn{10}{c}{\textit{Open-source VLMs}} \\
\midrule
Qwen2.5-VL-Instruct
& 7B
& 83.8 & 36.1 & 41.1 & 37.8 
& 63.9 & 37.1 & 26.3 & 29.6 \\
MiMO-VL-RL
& 7B
& 73.7 & 45.3 & 43.3 & 42.2 
& 48.6 & 35.2 & 28.8 & 30.6 \\
InternVL3 
& 8B
& 84.5 & 36.5 & 42.1 & 38.5 
& 73.4 & 48.8 & 42.6 & 44.2 \\
Ovis2.5
& 9B
& 80.1 & 30.7 & 32.6 & 31.0 
& 58.3 & 30.0 & 21.4 & 23.8 \\
Eagle2
& 9B
& 86.1 & 41.3 & 44.3 & 41.7 
& 64.1 & 38.4 & 29.0 & 31.7  \\
Qwen2.5-VL-Instruct
& 72B
& 95.5 & \textbf{47.5} & 42.3 & 43.8 
& 76.0 & 50.7 & 46.9 & 47.5 \\

\midrule
\multicolumn{10}{c}{\textit{Visual Reasoning Frameworks}} \\
\midrule
Pixel-Reasoner 
& --
& 75.2 & 28.3 & 40.6 & 32.6 
& 69.0 & 26.4 & 26.2 & 25.0 \\
Chain-of-Region 
& --
& \textbf{95.6} & 36.0 & 35.3 & 35.0 
& 82.5 & 57.4 & 43.2 & 49.3 \\
Speculative Verdict
& -- 
& 93.6 & 41.4 & 46.2 & 42.8 
& 79.4 & 47.4 & 37.8 & 40.7 \\

\midrule
\textbf{TopoAgent (Ours)} 
& --
& \textbf{95.6} & 43.9 & \textbf{49.9} & \textbf{45.8}
& \textbf{87.3 } & \textbf{63.0} & \textbf{56.8} & \textbf{58.1} \\

\bottomrule
\end{tabular*}
\caption{Main results on TopoBench-180. We report Node F1 and directed Edge Precision (P), Recall (R), and F1 on Web-style and Network-style diagrams. Model sizes are shown when available. Best results are highlighted in bold.}
\label{tab:main_results}
\end{table*}

\section{Experiments}
\subsection{Experimental Setup}
\paragraph{Baselines.}
We compare TopoAgent with representative vision-language models (VLMs) and visual reasoning frameworks. Our baselines include two strong closed-source VLMs, GPT-4o and GPT-4o-mini~\cite{hurst2024gpt}; a diverse set of open-source VLMs, including Qwen2.5-VL-Instruct (7B/72B)~\cite{wang2024qwen2}, MiMO-VL-RL (7B)~\cite{xiaomi2025mimo}, InternVL3 (8B)~\cite{zhu2025internvl3}, Eagle 2 (9B)~\cite{li2025eagle}, and Ovis2.5 (9B)~\cite{lu2025ovis2}; and recent visual reasoning frameworks for information-dense images, including Pixel-Reasoner~\cite{wang2025pixel}, Chain-of-Region~\cite{li2025chain}, and Speculative Verdict~\cite{liu2025small}. This baseline set covers both frontier general-purpose VLMs and recent structured reasoning systems for complex visual understanding. 

\paragraph{Dataset.}
We evaluate our method on TopoBench-180, the human-verified benchmark introduced in Section~\ref{sec:dataset}. We report results on two subsets: Web-style diagrams (54 images), where relations are mainly expressed by directed arrows, and Network-style diagrams (126 images), where connections are typically bidirectional or undirected. This split allows us to evaluate both directed relation extraction and symmetric connectivity reasoning.

\paragraph{Evaluation Metrics.}
We evaluate node extraction and relational structure extraction separately. For nodes, we report Node F1. For relations, we report Edge Precision, Edge Recall, and Edge F1. A predicted edge is counted as correct only if its $(source, target)$ tuple exactly matches a ground-truth edge within the corresponding node's relationship group. All metrics are computed after label normalization to reduce minor textual variations. More implementation details are provided in the Appendix.

% \paragraph{Implementation Details.}
% All agents in our framework are implemented using GPT-4o. Each stage invokes the model with a task-specific prompt and deterministic decoding (temperature $=0.0$). Intermediate raw responses and parsed JSON artifacts are persisted per stage for traceability and failure analysis.
\begin{table*}[t]
\centering
\small
\setlength{\tabcolsep}{6pt}
\begin{tabular}{l cccc|cccc}
\toprule
 & \multicolumn{4}{c|}{Web-style} 
 & \multicolumn{4}{c}{Network-style} \\
\cmidrule(lr){2-5} \cmidrule(lr){6-9}
Variant 
& Node F1 & Edge P & Edge R & Edge F1
& Node F1 & Edge P & Edge R & Edge F1 \\
\midrule

w/o Region Summary
& 87.2 & 43.9 & 41.5 & 40.8 
& 79.8 & 61.1 & 45.5 & 50.4 \\

w/o Figure Classification
& 95.3 & 41.0 & 49.7 & 45.6 
& 84.7 & 57.0 & 51.8 & 52.6 \\

w/o Node-centric Reasoning
& 95.7 & 43.2 & 37.2 & 39.4 
& 86.6 & 59.7 & 44.6 & 49.3 \\

w/o Consistency Enforcement
& 86.6 & 36.9 & 44.6 & 38.9 
& 81.7 & 54.3 & 49.5 & 50.7 \\

\midrule
TopoAgent (Full)
& 95.6 & 43.9 & 49.9 & 45.8
& 87.3 & 63.0 & 56.8 & 58.1 \\

\bottomrule
\end{tabular}
\caption{Ablation study of TopoAgent on TopoBench-180. We evaluate the contribution of each major component, including region-level perceptual summarization, global figure attribute inference, node-centric local-to-global reasoning, and final topological consistency enforcement. We report Node F1 and directed Edge Precision (P), Recall (R), and F1 on the Web-style and Network-style subsets.}
\label{tab:ablation_main}
\end{table*}

\subsection{Results on TopoBench-180.}

Table~\ref{tab:main_results} reports the main results on TopoBench-180. Overall, TopoAgent achieves the strongest topology extraction performance across both subsets, with the clearest gains appearing on edge-level metrics. This suggests that the proposed perception-to-reasoning framework is particularly effective for relational structure extraction, especially in settings where edge directionality or topology consistency is difficult to resolve from local cues alone. Across all methods, node extraction is generally easier than directed edge recovery, indicating that the main challenge lies in constructing correct relational topology rather than identifying isolated entities. 

On Web-style diagrams, TopoAgent achieves 95.6 Node F1 and 45.8 Edge F1, matching the best node extraction performance while outperforming all baselines in overall relation recovery. Compared with the strongest open-source VLM baseline, Qwen2.5-VL-Instruct (72B), it improves Edge F1 from 43.8 to 45.8, and compared with the strongest visual reasoning baseline, Speculative Verdict, it improves Edge F1 from 42.8 to 45.8. Although Qwen2.5-VL-Instruct (72B) attains the highest Edge Precision (47.5), TopoAgent achieves the highest Edge Recall (49.9), leading to the best overall Edge F1. This pattern suggests that Web-style diagrams often provide visually distinctive nodes but direction-sensitive edges, making directionality a key source of hallucination and edge inconsistency for existing models. The relatively smaller margin on this subset may also reflect that Web-style diagrams are closer to scientific visual question answering and diagram understanding data already seen during large-scale VLM pretraining or evaluation.

The gains are more pronounced on the Network-style subset, which places greater demands on topology-level consistency reasoning. Here, TopoAgent achieves 87.3 Node F1, 63.0 Edge Precision, 56.8 Edge Recall, and 58.1 Edge F1, outperforming all baselines. In particular, it surpasses GPT-4o by 4.6 Edge F1 points (58.1 vs.\ 53.5), Chain-of-Region by 8.8 points (58.1 vs.\ 49.3), and Qwen2.5-VL-Instruct (72B) by 10.6 points (58.1 vs.\ 47.5). Unlike on Web-style diagrams, TopoAgent achieves the highest precision and recall simultaneously on this subset. We attribute this larger gain to the stronger role of topological consistency enforcement: when connections are predominantly bidirectional or undirected, explicit structural constraints become especially effective for resolving ambiguous connectivity.

Overall, these results show that the combination of perception \& grounding and topology reasoning leads to more accurate and robust topology extraction, addressing direction-sensitive errors in Web-style diagrams and yielding especially clear gains on Network-style diagrams.

% \begin{table*}[t]
% \centering
% \small
% \setlength{\tabcolsep}{6pt}
% \begin{tabular}{l cccc|cccc}
% \toprule
%  & \multicolumn{4}{c|}{Web-style (54)} 
%  & \multicolumn{4}{c}{Network (126)} \\
% \cmidrule(lr){2-5} \cmidrule(lr){6-9}
% Variant 
% & Node F1 & Edge P & Edge R & Edge F1
% & Node F1 & Edge P & Edge R & Edge F1 \\
% \midrule

% w/o Region Summary
% & -- & -- & -- & -- 
% & -- & -- & -- & -- \\

% w/o Figure Classification
% & -- & -- & -- & -- 
% & -- & -- & -- & -- \\

% w/o Node-centric Reasoning
% & -- & -- & -- & -- 
% & -- & -- & -- & -- \\

% w/o Self-Check
% & -- & -- & -- & -- 
% & -- & -- & -- & -- \\

% \midrule
% \textbf{TopoAgent (Full)}
% & \textbf{95.6} & \textbf{43.9} & \textbf{49.9} & \textbf{45.8}
% & \textbf{87.3} & \textbf{61.2} & \textbf{55.3} & \textbf{56.1} \\

% \bottomrule
% \end{tabular}
% \caption{Ablation study of TopoAgent on TopGraph-180. We evaluate the contribution of each major module, including region-level visual summarization, figure-level structural classification, node-centric relation reasoning, and final self-check verification. We report Node F1 and directed Edge Precision (P), Recall (R), and F1 on the Web-style and Network subsets.}
% \label{tab:ablation_main}
% \end{table*}

\section{Ablation}
Table~\ref{tab:ablation_main} reports the ablation results on TopoBench-180. Overall, all major components contribute to the final performance, and removing any one of them leads to a drop in Edge F1 on both subsets. Region-level perceptual summarization is particularly important for stable node grounding, as removing it causes substantial degradation in both Node F1 and Edge F1. This suggests that localized perceptual evidence remains necessary even for strong vision-language models. Figure-level attribute inference provides useful global priors, but its contribution is smaller than that of the other modules, consistent with its role as a coarse structural guide rather than a direct substitute for fine-grained grounding or relation reasoning. Nevertheless, its consistent contribution across both subsets indicates that even coarse figure-level priors help narrow the reasoning space for downstream edge prediction.

The largest performance drops occur when removing node-centric reasoning or the final self-check stage. Without node-centric local-to-global reasoning, Edge Recall declines sharply on both subsets, especially for Web-style diagrams, highlighting the importance of node-anchored decomposition for recovering direction-sensitive and spatially extended relations. Removing self-check also reduces both Node F1 and Edge F1, showing that consistency enforcement is crucial for filtering invalid predictions and improving graph-level coherence. In particular, the drop after removing self-check indicates that local predictions alone are often insufficient to guarantee globally valid topology.

Overall, these results confirm that TopoAgent benefits from the complementary effects of perception \& grounding and topology reasoning, with different modules contributing to local grounding, relation coverage, and final graph consistency.

\section{Conclusion}
We presented TopoAgent, a structure-aware perception-to-reasoning framework for diagram-to-graph topology extraction, together with TopoBench-180, a human-verified benchmark for this task. TopoAgent progressively extracts graph structure from complex diagrams through grounded perception, global structural priors, canonical node inventory construction, node-centric local-to-global relation reasoning, and topological consistency enforcement. Experiments on TopoBench-180 show that this staged, structure-aware design improves topology extraction over strong vision-language model baselines and recent visual reasoning frameworks, especially on edge extraction. We hope this work will support future research on structured diagram understanding and topology-aware visual reasoning.

\section{Limitations}

\paragraph{Benchmark coverage.}
Despite these contributions, several limitations remain. Diagram-to-graph benchmarks are still scarce, and TopoBench-180 does not yet cover the full diversity of structural diagrams found in real-world technical documents. Although the benchmark includes both Web-style and Network-style diagrams, its scale and coverage remain limited relative to the range of diagram formats encountered in practice. Future extensions could scale the benchmark using synthetic or generative diagrams followed by human verification, which may increase structural diversity while reducing annotation ambiguity and licensing constraints.

\paragraph{Annotation cost.}
Topology annotation is labor-intensive. Producing canonical graph annotations requires exhaustive identification of nodes, connectors, and directionality, together with careful human verification. This limits the speed at which larger and more diverse benchmarks can be constructed.

\paragraph{Model limitations.}
Our framework relies on prompting general-purpose vision-language models rather than task-specific training. Although this design provides flexibility and reduces the need for specialized supervision, it also inherits the weaknesses of current foundation models. In particular, hallucination errors may still occur when inferred relations are not fully grounded in the image, and failures in fine-grained perception can lead to missing or incorrect edges.

\paragraph{Efficiency and scalability.}
TopoAgent adopts a multi-stage perception-to-reasoning pipeline, which improves extraction quality but also increases inference complexity relative to one-shot prediction. In practice, this design may introduce additional latency and computational cost, especially for larger or more visually complex diagrams. Future work may improve efficiency through more compact reasoning pipelines, stronger diagram-specific perception modules, and lightweight fine-tuning strategies that reduce reliance on repeated multi-stage inference.

\bibliography{references}

\newpage
\appendix

\section{Full Task Definition and Output Format}
\label{app:task}

\subsection{Task Definition}
Given a structural diagram image $I$, the goal of diagram-to-graph topology extraction is to predict a canonical graph $G=(V,E)$ that represents the entities in the diagram and their visual connections. Each node $v \in V$ corresponds to a valid diagram entity, typically identified by a textual label and associated visual region, and each edge $e \in E$ denotes a visible relation indicated by connectors such as arrows or lines.

Formally, the task is to learn a mapping
\[
f: I \rightarrow G=(V,E),
\]
where the predicted graph is compared against a canonical gold annotation. The node set $V$ contains all valid entities in the topology, and the edge set $E$ contains all visible relations among them. Depending on the diagram type, edges may be directed, bidirectional, or undirected. For evaluation, nodes and edges are normalized into a structured output format.

To make evaluation deterministic, each diagram is annotated using a canonical JSON schema, shown in Listing~\ref{lst:json_schema}. The top-level keys correspond to valid node names, and \textbf{each node stores only the outgoing relationships originating from that node}. Each relationship entry specifies the source node, the target node, and the relation type, so that the full graph can be extracted by aggregating the outgoing relations of all nodes.

\begin{lstlisting}[language=json,caption={Canonical annotation schema.},label={lst:json_schema}]
{
  "<node_name>": {
    "relations": [
      {
        "source": "<node_name>",
        "target": "<target_node>",
        "type": "<relation_type>"
      },
      ...
    ]
  },
  ...
}
\end{lstlisting}

\subsection{Illustrative Examples}
Figure~\ref{fig:annotation_example} shows a Web-style example of diagram-to-graph topology extraction together with its canonical JSON annotation. A corresponding Network-style example is provided below.

\begin{center}
    \includegraphics[width=0.9\linewidth]{figures/figure4.pdf}
    \captionof{figure}{A toy example of a Web-style diagram and its corresponding JSON annotation.}
    \label{fig:annotation_example}
\end{center}

\begin{tcolorbox}[
    title=Example JSON Annotation,
    colback=gray!5,
    colframe=black!50,
    boxrule=0.5pt,
    arc=2pt,
    left=6pt,
    right=6pt,
    top=6pt,
    bottom=6pt
]
\begin{lstlisting}[language=json,caption={Example canonical JSON annotation.},label={lst:json_example}]
{
  "Berries": {
    "relations": [
      {
        "source": "Berries",
        "target": "Greenfly",
        "type": "directed"
      },
      {
        "source": "Berries",
        "target": "Butterfly",
        "type": "directed"
      }
    ]
  },
  "Plantain": {
    "relations": [
      {
        "source": "Plantain",
        "target": "Grasshopper",
        "type": "directed"
      },
      {
        "source": "Plantain",
        "target": "Mouse",
        "type": "directed"
      },
      {
        "source": "Plantain",
        "target": "Rabbit",
        "type": "directed"
      }
    ]
  },
  ...
  "Buzzard": {
    "relations": []
  },
  "Fox": {
    "relations": []
  }
}
\end{lstlisting}
\end{tcolorbox}

Figure~\ref{fig:annotation_example2} shows a network-style example of diagram-to-graph topology extraction together with its canonical JSON annotation.

\begin{center}
    \includegraphics[width=\linewidth]{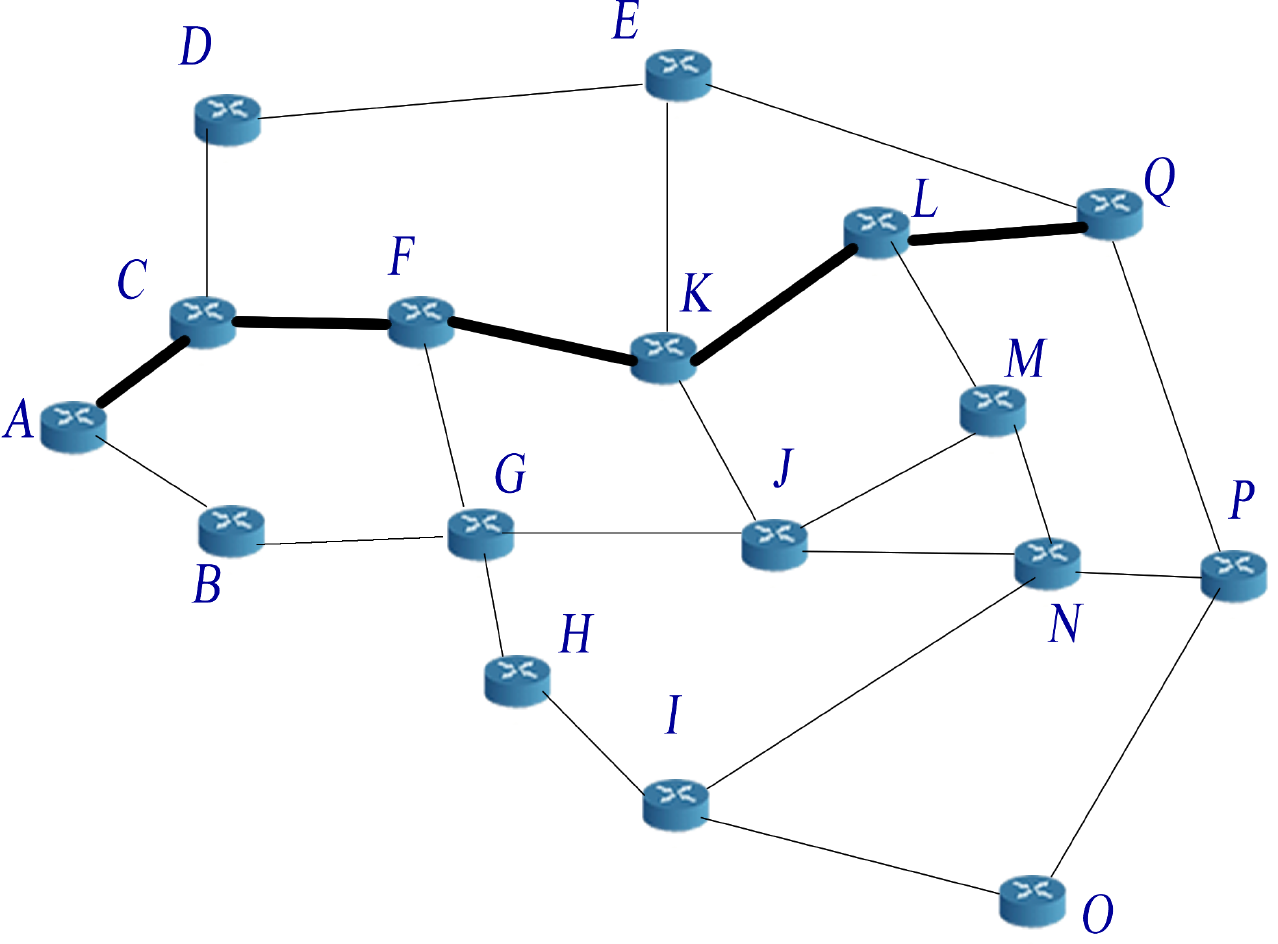}
    \captionof{figure}{A toy example of a network-style diagram and its corresponding graph annotation.}
    \label{fig:annotation_example2}
\end{center}

\begin{tcolorbox}[
    title=Example JSON Annotation,
    colback=gray!5,
    colframe=black!50,
    boxrule=0.5pt,
    arc=2pt,
    left=6pt,
    right=6pt,
    top=6pt,
    bottom=6pt
]
\begin{lstlisting}[language=json]
{
  "A": {
    "relations": [
      {
        "source": "A",
        "target": "B",
        "type": "undirected"
      },
      {
        "source": "A",
        "target": "C",
        "type": "undirected"
      }
    ]
  },
  "C": {
    "relations": [
      {
        "source": "C",
        "target": "A",
        "type": "undirected"
      },
      {
        "source": "C",
        "target": "D",
        "type": "undirected"
      }
      {
        "source": "C",
        "target": "F",
        "type": "undirected"
      },
    ]
  },
  ...
}
\end{lstlisting}
\end{tcolorbox}

\section{Benchmark Construction Details}
\label{app:benchmark}

\subsection{Data Sources}

\paragraph{Google-based candidate retrieval.}
As one source of candidate diagrams, we queried Google Images through SerpAPI using the following topology-related search terms:
\textit{network topology diagram},
\textit{computer network topology diagram},
\textit{communication network diagram},
\textit{network architecture diagram},
\textit{network layout diagram},
\textit{internet backbone topology diagram},
\textit{data center network topology diagram},
\textit{enterprise network topology diagram},
\textit{wireless network topology diagram},
\textit{router switch topology diagram},
\textit{lan wan network diagram},
\textit{sdn network topology diagram},
\textit{server network diagram},
\textit{system connectivity diagram},
\textit{infrastructure topology diagram}, and
\textit{distributed system network diagram}.

For each query, we retrieved results from two pages and downloaded the returned images together with their metadata. For every retrieved candidate, we recorded the image URL, source page URL, title, query provenance, and content type, and retained both raster images and SVG files whenever available. We then applied file-level validation to remove clearly invalid candidates, discarding files with non-200 HTTP status codes, HTML content, extremely small file size, or unreadable image content. Duplicate candidates were removed through both URL-level matching and content-level deduplication using SHA-1 hashes. To maximize recall at the crawling stage, we adopted permissive retention rules and kept potentially valid files for later inspection whenever possible, while deferring stricter semantic filtering to subsequent manual verification. This Google-based retrieval stage served as only one component of the broader collection pipeline; additional candidate diagrams were gathered from other public sources in subsequent stages.

\paragraph{Additional public sources.}
To complement Google-based retrieval, we further collected candidate diagrams from several public sources, including Wikipedia pages, Wikimedia Commons categories, OpenClipart, SVGRepo, and arXiv. For Wikipedia and Wikimedia Commons, we queried pages and categories related to topology diagrams, block diagrams, graph structures, circuit schematics, wiring diagrams, hydraulic diagrams, and other engineering-style visual structures. We extracted linked image files, retrieved their full-resolution URLs through the Wikimedia API, and retained raster images and SVG files that satisfied basic validity and size constraints. To improve relevance, we applied lightweight filename-based filtering using topology- and diagram-related keywords, while excluding obvious non-diagram assets such as plots, benchmark figures, and natural images.

For OpenClipart and SVGRepo, we issued diagram-related search queries such as \textit{network diagram}, \textit{topology diagram}, \textit{flowchart}, \textit{block diagram}, and \textit{circuit diagram}. We retained only assets with permissive licenses when applicable and applied the same lightweight filtering rules based on filenames, extensions, and basic file validity. Since these sources often provide vector graphics, SVG files were preserved whenever available.

For arXiv, we queried papers using diagram-related terms such as \textit{flowchart}, \textit{block diagram}, \textit{network topology diagram}, \textit{node-link diagram}, \textit{circuit schematic}, and \textit{control system diagram}. For each matched paper, we downloaded the source tarball and extracted embedded raster images from the source files. We retained only images that satisfied minimum size constraints and excluded files whose names suggested plots, tables, benchmark figures, or other non-diagram content. Together, these additional public sources expanded the diversity of candidate diagrams beyond web image retrieval and improved coverage of structurally rich technical diagrams from educational, engineering, and scientific documents.

\paragraph{Source composition.}
Before filtering and annotation, the candidate pool contains over 6,000 raw images from both public and internal sources. Among the public sources, 3,088 images come from AI2D~\cite{kembhavi2016diagram}, 1,605 from Google image search, and 1,627 from Wikipedia- and arXiv-based retrieval, and 78 from the Network Topology Diagram Computer Vision Dataset\footnote{\url{https://universe.roboflow.com/project-v2jjh/network-topology-diagram}}. In addition, we collected 17 communication network design diagrams internally to supplement engineering-style topology layouts that are less common in public resources. This source composition improves diversity in diagram domain, layout style, and connectivity structure.

\subsection{Automatic Filtering and Inclusion Criteria}

After candidate retrieval, we applied a lightweight filtering stage to remove clearly invalid or low-value samples before manual review. First, we discarded files that were not valid images, including HTML pages, corrupted files, or failed downloads. We further removed exact duplicates through URL-level matching and content-level deduplication. To reduce obviously irrelevant samples, we applied simple text-based filtering using filenames, titles, and source metadata, retaining candidates whose associated text suggested diagram- or topology-related content and excluding files with indicators of plots, benchmark figures, tables, or natural photographs.

We also filtered out very small images that were unlikely to support reliable topology annotation. In particular, any raster image with width or height smaller than 128 pixels was removed. This step helped eliminate low-resolution thumbnails, icons, and other visually insufficient assets while preserving structurally informative diagrams. After this step, the initial pool of over 6,000 raw diagrams was reduced to 4,318 unique candidates.

\subsection{VLM-driven pruning}
After lightweight automatic filtering, we further applied a two-stage GPT-driven pruning process to improve candidate quality before final manual verification. We use the GPT-4o API to perform the filtering step. The first stage was designed to be recall-oriented, removing only clearly irrelevant samples while preserving a broad range of potentially valid diagrams. The second stage was more strict and precision-oriented, aiming to retain only structurally valid diagrams that were suitable for canonical graph annotation.

\paragraph{Stage 1: Recall-oriented coarse pruning.}
In the first stage, GPT was used as a semantic triage model to distinguish likely diagram candidates from obviously out-of-scope images. This stage accepted a broad range of structural diagrams, including topology diagrams, node--edge graphs, block diagrams, flowcharts, pipelines, system architectures, and schematic or wiring-style diagrams. At the same time, it rejected clearly irrelevant samples such as natural photographs, screenshots, tables, charts, scanned documents, text-dominant pages, and hand-drawn sketches. Because the goal of this stage was to maximize recall, the filtering criterion was intentionally permissive, so that potentially useful structural diagrams would not be removed prematurely. Listing~\ref{lst:stage1_prompt} shows the prompt used in this stage.

\begin{figure*}[t]
\centering
\begin{minipage}{0.96\textwidth}
\begin{lstlisting}[
    caption={Prompt used for Stage 1 GPT filtering.},
    label={lst:stage1_prompt},
    basicstyle=\ttfamily\footnotesize,
    breaklines=true,
    breakatwhitespace=false,
    columns=fullflexible,
    keepspaces=true,
    frame=single
]
You are an image triage assistant.
Decide if the image is a diagram/architecture/topology/flow/block/schematic.

Say YES if it is any of:
- network topology / node-edge graph / boxes-and-arrows / block diagram / flowchart / pipeline
- system architecture, data flow, components/modules with arrows/lines
- circuit/wiring/schematic symbols, instrumentation/ladder diagrams

Say NO if it is any of:
- real-world photos (people/objects/scenes), UI/screenshots, tables
- charts/plots (line, bar, pie, scatter, box/violin, histogram, heatmap, confusion matrix, ROC/PR)
- scanned documents or mostly text
- hand-drawn or hand-written diagrams, sketches, whiteboard photos

Reply with a single word: YES or NO.
\end{lstlisting}
\end{minipage}
\end{figure*}
\paragraph{Stage 2: Precision-oriented structural filtering.}
In the second stage, we applied a stricter GPT-based filtering step to assess whether a candidate diagram was suitable for benchmark annotation under our task definition. Unlike the first-stage semantic triage, which aimed to preserve broad structural coverage, this stage focused on annotation feasibility and structural validity. Specifically, the model was asked to inspect the candidate diagram, identify readable node labels, estimate the number of explicit connections, assess visual clarity, and determine whether the overall structural complexity was appropriate for reliable canonical graph annotation. A candidate was retained only if it contained between 4 and 30 identifiable nodes, exhibited explicit and readable relations, and was neither trivially simple nor excessively complex for consistent transcription. We also required the diagram to have sufficient connectivity, with the estimated number of explicit relations being at least the number of identified nodes minus one. This stage therefore served as a precision-oriented filter before final human verification. After the two-stage VLM-driven pruning process, the candidate pool was reduced to 741 structurally valid diagrams. Listing~\ref{lst:stage2_prompt} shows the prompt used in this stage.

% \begin{figure}[!hb]
%     \centering
%     \includegraphics[width=\linewidth]{figures/figure8.pdf}
%     \caption{
%     Complexity analysis of TopoBench-180. We report node count, relation count, and unique-pair density for Web-style and Network-style diagrams. The distributions show that diagram difficulty is not determined by node count alone, but also by edge density and structural connectivity.
%     }
%     \label{fig:complexity}
% \end{figure}

\begin{figure*}[t]
\centering
\begin{minipage}{0.96\textwidth}
\begin{lstlisting}[
    caption={Prompt used for Stage 2 GPT filtering.},
    label={lst:stage2_prompt},
    basicstyle=\ttfamily\footnotesize,
    breaklines=true,
    breakatwhitespace=false,
    columns=fullflexible,
    keepspaces=true,
    frame=single
]
You are a strict structural diagram screening assistant for benchmark construction.

Your goal is to determine whether the image is suitable for inclusion in a diagram-to-graph topology extraction benchmark.

Evaluate the image according to the following annotation suitability criteria:
- node_names: identify all valid entities or nodes with readable labels.
- edge_count: estimate the number of explicit visible relations between nodes, indicated by arrows, lines, or equivalent link primitives.
- clarity: assign "good" if node labels are readable and connections are visually clear enough for reliable graph annotation; otherwise assign "bad".
- complexity:
  - "too_simple" if the image contains 3 or fewer valid nodes,
  - "ok" if the diagram is suitable for canonical graph annotation,
  - "too_complex" if the structure is too dense or visually overwhelming for reliable transcription.

Decision rule:
Return "YES" only if:
1. the number of identified nodes is between MIN_NODES and MAX_NODES,
2. clarity is "good",
3. edge_count >= number_of_nodes - 1.

Otherwise return "NO".

Return JSON only in the following format:
{"node_names": [...], "edge_count": <int>, "clarity": "good|bad", "complexity": "too_simple|ok|too_complex", "decision": "YES|NO"}

If any part of the image is ambiguous or uncertain, return "decision": "NO".
\end{lstlisting}
\end{minipage}
\end{figure*}

\subsection{Human-in-the-Loop Verification}
After the two-stage VLM-driven pruning process, the remaining candidates were manually reviewed by two annotators. Each annotator re-examined the retained diagrams to verify whether the nodes and relations were visually identifiable and whether the image was suitable for reliable canonical graph annotation. Diagrams that were visually ambiguous, blurry, or insufficiently clear for consistent transcription were removed at this stage. In addition, annotators considered the GPT outputs from the second-stage structural filtering when excluding candidates that were too simple or too complex for the target benchmark setting. After this final human verification and pruning step, we retained 180 diagrams for benchmark annotation and evaluation.

\subsection{Gold-Standard Graph Canonicalization}

\paragraph{VLM Draft Graph Generation.}
To reduce the annotation burden, we first use GPT-4o to generate a draft graph for each candidate diagram. This draft graph is intended only as an initial structural proposal, including candidate nodes and relations, and is used to accelerate the subsequent manual annotation process. At this stage, the VLM output is not treated as a gold annotation and is not assumed to be correct.

\paragraph{Expert Topology Calibration.}
Human annotators then review and calibrate each draft graph against the original diagram to produce the final gold-standard annotation. During this process, annotators verify all nodes, relations, and directionality, remove hallucinated or invalid elements, add missing nodes or edges, and normalize the graph into the canonical output format. Node labels are preserved exactly as they appear in the original diagram, without translation or language normalization. When a legend is present, annotators follow its explicitly stated semantics when interpreting node and relation types. The draft graph serves only as an annotation aid to reduce manual effort; all final annotations are established through human verification and correction. As a result, the released gold-standard graphs reflect expert-calibrated topology rather than raw VLM output. Dataset construction and annotation were completed independently of downstream model evaluation, and no evaluation metrics were used during candidate filtering or annotation refinement.

\paragraph{Node Disambiguation.}
For nodes with repeated textual labels, we apply a deterministic naming convention during canonicalization to ensure unique identification. Specifically, repeated nodes are indexed by spatial order from left to right and top to bottom, using the form \texttt{<NODE\_NAME>\_<index>}. If a repeated node appears inside a larger named enclosure, the enclosure is also indexed and the node is recorded as \texttt{<NODE\_NAME>\_<index>\_<enclosure>}. This rule is introduced solely to resolve naming ambiguity in the structured output format.

\subsection{Dataset Complexity Analysis}

To better characterize the complexity of TopoBench-180, we provide a quantitative analysis of its structural properties. Table~\ref{tab:complexity_stats} summarizes the minimum, mean, and maximum node count, relation count, and unique-pair density for the Web-style and Network-style subsets, while Figure~\ref{fig:complexity} shows their full distributions. The unique-pair density is computed as the ratio between the number of directed relations and the maximum possible number of directed node pairs, i.e., $|E|/(|V|(|V|-1))$.

\begin{table}[t]
\centering
\small
\resizebox{0.82\linewidth}{!}{
\begin{tabular}{llccc}
\toprule
\textbf{Dataset} & \textbf{Metric} & \textbf{Min} & \textbf{Mean} & \textbf{Max} \\
\midrule
Web-style & Nodes   & 5     & 10.09 & 23    \\
 & Edges   & 5     & 14.35 & 37    \\
 & Density & 0.055 & 0.174 & 0.450 \\
\midrule
Network-style & Nodes   & 4     & 11.56 & 37    \\
 & Edges   & 5     & 23.09 & 79    \\
 & Density & 0.037 & 0.230 & 1.000 \\
\bottomrule
\end{tabular}
}
\caption{
Summary statistics of structural complexity in TopoBench-180. We report the minimum, mean, and maximum node count, edge count, and graph density for the Web-style and Network-style subsets.
}
\label{tab:complexity_stats}
\end{table}

The two subsets exhibit different sources of difficulty. Web-style diagrams generally contain fewer nodes and relations and have lower density, but they remain sensitive to relation direction, local layout cues, and textual semantics. In contrast, Network-style diagrams span a wider range of node and relation counts and exhibit higher density on average, making topology completeness and consistent node grounding more challenging.

\begin{center}
    \includegraphics[width=\linewidth]{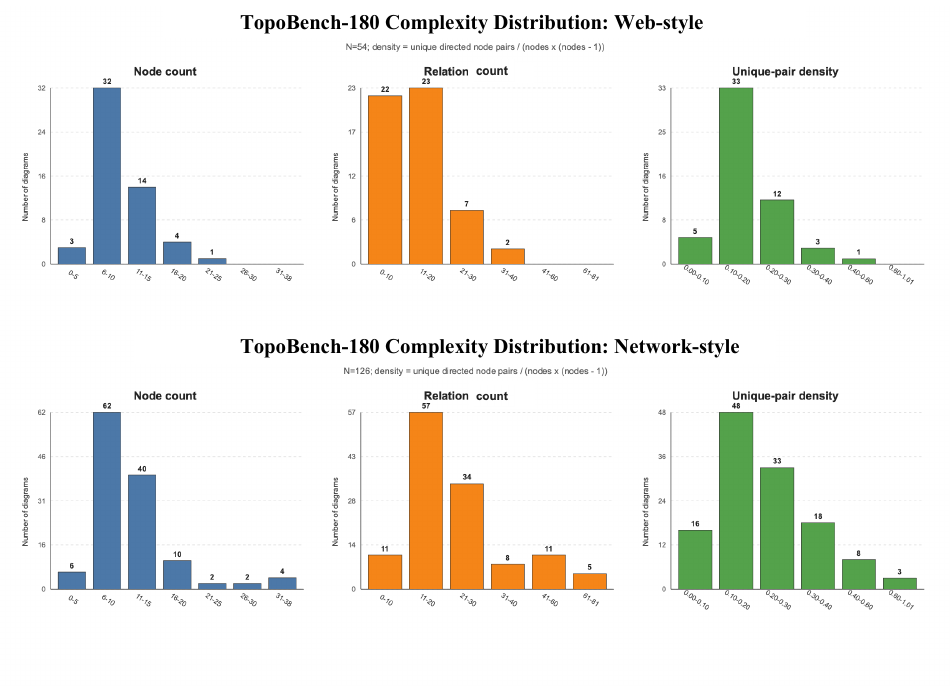}
    \captionof{figure}{
    Complexity analysis of TopoBench-180. We report node count, relation count, and unique-pair density for Web-style and Network-style diagrams. The distributions show that diagram difficulty is not determined by node count alone, but also by edge density and structural connectivity.
    }
    \label{fig:complexity}
\end{center}
% \begin{figure}[!htbp]
%     \centering
%     \includegraphics[width=\linewidth]{figures/figure8.pdf}
%     \caption{
%     Complexity analysis of TopoBench-180. We report node count, relation count, and unique-pair density for Web-style and Network-style diagrams. The distributions show that diagram difficulty is not determined by node count alone, but also by edge density and structural connectivity.
%     }
%     \label{fig:complexity}
% \end{figure}

\section{Implementation Details of TopoAgent}
\label{app:method}
\subsection{Perception and OCR Grounding}
\label{appendix:perception_ocr_grounding}

Before topology reasoning, we perform a lightweight perception stage to decompose the input diagram into localized visual regions and associate them with grounded textual evidence. This stage follows a chain-of-region~\cite{li2025chain} style pipeline, where the image is progressively transformed into a set of region-level visual-textual units that can support downstream node inventory extraction, region summarization, and relation reasoning.

\paragraph{Overview.}
Let $X_{\mathrm{im}} \in \mathbb{R}^{H \times W \times 3}$ denote the input diagram image. Instead of relying on a single-pass holistic interpretation, we first partition the image into smaller visual regions, refine overly coarse connected components, and merge fragmented regions when necessary. The final regions are then paired with OCR outputs to form grounded perceptual units. This process improves locality and reduces the difficulty of directly reasoning over dense, cluttered diagrams.

\paragraph{Region initialization.}
We begin by constructing an initial region map using connected-component analysis on a binarized version of the image. Specifically, the image is converted into a binary map $X_{\mathrm{bi}}$ using Otsu thresholding, after which connected components are extracted from both the foreground and its inverse:
\begin{align}
X_{\mathrm{bi}} &= \mathrm{Otsu}(X_{\mathrm{im}}), \\
X_{\mathrm{fg}} &= \mathrm{CC}(X_{\mathrm{bi}}), \\
X_{\mathrm{bg}} &= \mathrm{CC}(1 - X_{\mathrm{bi}}),
\end{align}
where $\mathrm{CC}(\cdot)$ denotes connected-component labeling under 8-connectivity. The two label maps are then merged with an offset to avoid label collisions:
\begin{equation}
X_{\mathrm{region}} = X_{\mathrm{fg}} + X_{\mathrm{bg}} + \mathbb{I}[X_{\mathrm{bg}} > 0]\cdot \mathrm{offset}.
\end{equation}
This initialization captures both dark-on-light and light-on-dark visual structures and provides a coarse decomposition of the diagram into candidate regions.

\paragraph{Region splitting.}
The initialized regions may still contain multiple connected visual elements, such as arrows attached to nodes or composite shapes with narrow bridges. To improve separation, we refine such regions through region splitting.

Given a region mask $M$ and a detected subregion mask $M_{\mathrm{sub}}$, we subtract the subregion and relabel the remaining connected parts:
\begin{align}
M_{\mathrm{remain}} &= M - M_{\mathrm{sub}}, \\
X_{\mathrm{remain}} &= \mathrm{CC}(M_{\mathrm{remain}}),
\end{align}
after which the global region map is updated accordingly.

We consider two splitting modes. For structured split, we apply specialized detectors for common geometric primitives such as rectangles, ellipses, and line segments. For a region mask $M_i$, the detector outputs a structured subregion $M_{\mathrm{shape}}$ together with optional geometric parameters:
\begin{equation}
(M_{\mathrm{shape}}, S_i) = \mathrm{StructureDetector}(M_i).
\end{equation}
The detected structure is then separated from the original region using the splitting procedure above.

For unstructured split, we adopt a connectivity-based strategy to break weakly connected composite shapes. Specifically, we compute a distance transform on the region mask and derive a scale parameter
\begin{equation}
T = \left\lfloor \frac{\max(\mathrm{DT}(M_i))}{2} \right\rfloor,
\end{equation}
where $\mathrm{DT}(\cdot)$ denotes the Euclidean distance transform. An elliptical structuring element of size $(T,T)$ is then used in a morphological opening operation:
\begin{equation}
M_{\mathrm{shape}} = \mathrm{Open}(M_i; \mathrm{EllipseKernel}(T,T)).
\end{equation}
This operation tends to preserve stable interior components while removing narrow connections, thereby producing more separable subregions.

\paragraph{Region merging.}
After splitting, the diagram may contain many small regions, which is undesirable for downstream processing. We therefore merge regions to obtain a more compact set of perceptual units. We first apply structured merging using OCR boxes and simple visual heuristics. Text regions detected by OCR are grouped with nearby or enclosed visual content, and visually regular auxiliary elements such as dotted lines or background lines can also be merged according to predefined rules. We then apply unstructured merging through agglomerative clustering over region masks. Let $M_i$ and $M_j$ denote two region masks and let $D_{ij}$ be the Euclidean distance between their centroids. Using this pairwise distance matrix, we iteratively merge regions until the total number of final regions reaches a predefined budget $B$. The resulting region set is denoted by $\{M_i^\ast\}_{i=1}^{B}$.

\paragraph{OCR grounding and region-level evidence construction.}
For each merged region $M_i^\ast$, we associate OCR text spans based on spatial overlap and proximity. This produces region-level visual-textual units that contain both localized image content and grounded text evidence. For structured regions, we also retain available geometric metadata, such as bounding boxes, contours, line endpoints, or ellipse parameters, when such information is available from the splitting stage.

\paragraph{Use in downstream reasoning.}
The final output of this stage is a set of grounded regions, each represented by its mask, optional structural attributes, and aligned OCR content. These region-level units are subsequently used to support region summarization, node extraction, and localized relation reasoning. In this way, the perception stage serves as an interface between low-level diagram appearance and higher-level topology inference.

\subsection{Region Summarization}

After region partition and OCR grounding, we summarize the visual content at both the local-region and global-figure levels to provide compact contextual evidence for downstream topology extraction.

\paragraph{Local region summarization.}
For each localized region, we query the VLM to produce a concise summary of only explicitly visible content. The summary is intended to capture grounded regional evidence that may support downstream node and relation extraction, such as node labels, legend content, and other local structural cues. To reduce hallucinated topology, the model is instructed to avoid inferring unseen relations. We use the following prompt:
\begin{quote}
\small
\textit{Summarize this local region for topology extraction. Focus on visible node labels, legend content and other structural cues. Only describe explicitly visible information. Do not infer missing relations.}
\end{quote}

\paragraph{Global figure summarization.}
In addition to local summaries, we obtain a high-level summary over the full diagram to capture global context that may not be recoverable from isolated regions alone. This summary focuses on visible node types and legend semantics, and serves as supplementary context for downstream graph extraction. The prompt is:
\begin{quote}
\small
\textit{You are given a full topology diagram. Summarize the figure at a high level for downstream graph extraction. Focus on visible node types and legend semantics.}
\end{quote}

\paragraph{Use in downstream extraction.}
The local summaries provide fine-grained grounded evidence, while the global summary contributes complementary high-level context. Together, they support later stages including figure classification, node inventory extraction, relationship reasoning, and final consistency checking.

\subsection{Local-to-Global Topology Reasoning}
\paragraph{Stage I: Global Figure Attribute Inference.} This stage predicts a set of global figure attributes,
\begin{equation}
A = \{a_{\text{sem}}, a_{\text{dir}}, a_{\text{dens}}\},
\end{equation}
which characterize the diagram semantics, link directionality, and link density, respectively. Concretely, given the full diagram together with auxiliary region summaries and preliminary node inventory, the model is prompted to classify the figure as \texttt{web}, \texttt{network}, or \texttt{structural\_diagram}; determine whether links are \texttt{one\_direction} or \texttt{bi\_direction}; and estimate whether inter-node connectivity is predominantly \texttt{single\_link} or \texttt{multi\_link}. Listing~\ref{lst:figure_classification_prompt} shows the prompt used in this stage.

\begin{figure*}[t]
\centering
\begin{minipage}{0.96\textwidth}
\begin{lstlisting}[
    caption={Prompt used for Stage I: Global Figure Attribute Inference.},
    label={lst:figure_classification_prompt},
    basicstyle=\ttfamily\footnotesize,
    breaklines=true,
    breakatwhitespace=false,
    columns=fullflexible,
    keepspaces=true,
    frame=single
]
You are performing global figure attribute inference for topology extraction.

Input:
- One full topology/network diagram image.
- Auxiliary pre-extracted region summaries and node inventory, which may be incomplete or imperfect.

Task:
Classify the figure along the following three dimensions based on visual evidence and the auxiliary context.

1. diagram_semantics
Return one of the following labels:
- "web": a heterogeneous explanatory or conceptual figure, often mixing multiple object types, text annotations, and irregular layouts.
- "network": a graph-like relational diagram where nodes are linked mainly through edge structures and the topology is the dominant visual organization.
- "structural_diagram": a diagram emphasizing parts, components, or functional structure, often with object-centric layouts rather than graph-centric connectivity.

2. link_directionality
- Return "one_direction" if most visible links contain explicit arrowheads or pointers indicating directed flow.
- Return "bi_direction" if links are mainly plain lines without arrowheads, explicitly two-way arrows, or if the same node pair appears in both directions.
- If both one-way and bidirectional patterns are present, return "bi_direction" and explain the bidirectional evidence.

3. link_density
- Return "multi_link" only if at least one specific node pair is connected by two or more visually distinct links.
- Otherwise return "single_link".

Requirements:
- Use the image as the primary evidence.
- Use the auxiliary region/node information only as supporting context.
- Provide one brief reason for each classification.
- If evidence is weak or ambiguous, choose the most conservative label supported by the image.

Return JSON only in the following format:
{
  "diagram_semantics": {
    "label": "web|network|structural_diagram",
    "reason": "<brief justification>"
  },
  "link_directionality": {
    "label": "one_direction|bi_direction",
    "reason": "<brief justification>"
  },
  "link_density": {
    "label": "multi_link|single_link",
    "reason": "<brief justification>"
  }
}

Do not output markdown, explanations, or any text outside the JSON object.
\end{lstlisting}
\end{minipage}
\end{figure*}

\paragraph{Stage II: Canonical Node Inventory Construction.}
The second stage constructs a canonical node inventory
\begin{equation}
V = \{v_i\}_{i=1}^{N},
\end{equation}
by consolidating visual evidence from the input image $I$, the OCR-grounded region-level summaries $\mathcal{S}$ into a stable symbolic node list. Each node corresponds to a semantically valid diagram entity grounded in readable textual labels and supporting visual evidence. In addition to the node set, this stage identifies discarded text regions that do not correspond to valid entities and summarizes legend information when present.

Given the full diagram, the regional OCR evidence, and the inferred figure profile, the model is prompted to produce a complete node inventory together with discarded non-entity labels and a legend summary. The resulting inventory serves as the canonical symbolic scaffold for downstream relationship reasoning and final consistency checking. Listing~\ref{lst:node_inventory_prompt} shows the prompt used in this stage.

\begin{figure*}[t]
\centering
\begin{minipage}{0.96\textwidth}
\begin{lstlisting}[
    caption={Prompt used for Stage II: Canonical Node Inventory Construction.},
    label={lst:node_inventory_prompt},
    basicstyle=\ttfamily\footnotesize,
    breaklines=true,
    breakatwhitespace=false,
    columns=fullflexible,
    keepspaces=true,
    frame=single
]
You are given a full topology/system/network diagram, auxiliary region-level OCR information, and a figure profile.

Task:
Construct a canonical inventory of all valid nodes in the diagram and summarize legend-defined link semantics.

Inputs:
1. Full diagram image.
2. Region information (pre-extracted text and location evidence), which may be incomplete or imperfect.
3. Figure profile JSON describing high-level diagram attributes.

Rules:
1. Identify all semantically valid connected entities using both the image and the region information.
2. If a labeled shape has neither visible links nor arrow-based references, exclude it from "nodes" and place it in "discarded" with a brief reason.
3. Preserve original labels and language exactly as shown. Do not translate or normalize wording.
4. By default, keep the original node label. If multiple nodes share the same label, append deterministic numeric suffixes using spatial order (left-to-right, then top-to-bottom).
5. Only in rare cases with clearly named enclosures containing multiple nodes, use enclosure-aware naming:
   <NodeLabel>_<NodeIndex>_<EnclosureName>_<EnclosureIndex>
6. Inspect the image and region information for any legend or visual key defining link semantics, and extract each relation type together with its text description and visual cues (e.g., color, line style, arrow style).

Return JSON only in the following format:
{
  "nodes": [
    {"name": "<final_node_name>", "evidence": "<short note on location or appearance>"}
  ],
  "discarded": [
    {"name": "<excluded_label>", "reason": "<why it was excluded>"}
  ],
  "legend_summary": [
    {"type": "<legend_name>", "color": "<visual cue>", "feature": "<functional description>"}
  ]
}

Do not output markdown, explanations, or any text outside the JSON object.
\end{lstlisting}
\end{minipage}
\end{figure*}

\paragraph{Stage III: Node-Centric Local-to-Global Relation Reasoning.}
The third stage performs relation reasoning in a node-centric manner. Instead of predicting all relations jointly under a single global visual context, it decomposes full-graph extraction into $N$ node-anchored subproblems, which is particularly helpful in dense diagrams where fine-grained connectors such as small arrows, thin lines, or crowded edge attachments may be missed under one-shot reasoning. For each focus node $v_i \in V$, the model takes the full diagram $I$, the grounded summaries $\mathcal{S}$, and the inferred global figure attributes $A$, and predicts only the relations associated with $v_i$. The resulting node-local edge set is denoted by
\begin{equation}
E_i = \mathcal{F}_{\text{rel}}(v_i, I, \mathcal{S}, A).
\end{equation}
Aggregating all node-local predictions yields a draft edge set
\begin{equation}
E_{\text{draft}} = \bigcup_{i=1}^{|V|} E_i .
\end{equation}

Given the canonical node inventory $V$, the selected focus node, the full diagram, and the legend-aware regional context, the model is prompted to reason about one node at a time and return only its associated local relations. This stage preserves global symbolic context while enforcing localized visual focus, thereby improving extraction of subtle connections in complex diagrams. Listing~\ref{lst:relationship_prompt} shows the prompt used in this stage.

\begin{figure*}[t]
\centering
\begin{minipage}{0.96\textwidth}
\begin{lstlisting}[
    caption={Prompt used for Stage III: Node-Centric Local-to-Global Relation Reasoning.},
    label={lst:relationship_prompt},
    basicstyle=\ttfamily\footnotesize,
    breaklines=true,
    breakatwhitespace=false,
    columns=fullflexible,
    keepspaces=true,
    frame=single
]
You are given a full topology diagram, a canonical node inventory with legend summary, one focus node, and inferred global figure attributes.

Task:
Identify all visible relationships associated with the focus node.

Inputs:
1. Focus node JSON.
2. Full node inventory and legend summary JSON.
3. Figure profile JSON.
4. Full diagram image.

Rules:
1. Return only relationships involving the focus node. Do not invent node names.
2. Use the image as primary evidence and the node/legend/profile information as supporting context.
3. Trace every visible connector associated with the focus node.
4. Determine relation types directly in this stage, using legend-defined terms when available.

Attribute-conditioned reasoning:
5. Use diagram_semantics as a structural prior:
   - "web": prefer directional interpretation when supported by arrows, flow, or semantic conventions.
   - "network": allow symmetric, bidirectional, or undirected interpretations when arrow evidence is weak or absent.
   - "structural_diagram": focus on component, attachment, or functional relations and infer direction conservatively.
   Always override the prior if the image provides stronger evidence.

6. Use link_directionality as a direction constraint:
   - "one_direction": prefer directed relations and follow arrowheads or pointers strictly.
   - "bi_direction": allow undirected or reciprocal interpretations when links lack clear directional evidence.
   - If direction is ambiguous, choose the most conservative relation supported by the image and record the ambiguity in "uncertain_links".

7. Use link_density to control parallel-edge extraction:
   - "single_link": return at most one relation for a node pair unless multiple distinct links are clearly visible.
   - "multi_link": explicitly check whether the same node pair is connected by multiple visually distinct links, and if so, return them separately.

Return JSON only in the following format:
{
  "nodes": {
    "<node_name>": {
      "relations": [
        {
          "source": "<node_name>",
          "target": "<target_name>",
          "type": "<connection_type_or_empty>"
        }
      ]
    }
  },
  "uncertain_links": [
    {
      "source": "<node>",
      "target": "<node>",
      "reason": "<ambiguity>"
    }
  ]
}

Do not output markdown, explanations, or any text outside the JSON object.
\end{lstlisting}
\end{minipage}
\end{figure*}

\paragraph{Stage IV: Topological Consistency Enforcement.}
The final stage performs topological consistency enforcement on the draft graph without revisiting the raw image, thereby reducing the risk of introducing new visual hallucinations during verification. Starting from the canonical node inventory $V$, the draft edge set $E_{\text{draft}}$, and the inferred global figure attributes $A$, this stage first applies an LLM-based self-check to consolidate node-local relation predictions into a unified graph representation, and then uses deterministic post-processing rules to enforce hard structural constraints. During the self-check step, the model is prompted to merge redundant local predictions, resolve directionality conflicts, and normalize the output into a single structured \texttt{JSON} object. In particular, the prompt encourages consistency with the inferred figure attributes, especially for edge directionality and reciprocal-link behavior. Listing~\ref{lst:self_check_prompt} shows the prompt used in this stage.

After the LLM self-check, we apply deterministic hard constraints as a final post-processing step rather than relying on the model alone. First, we enforce endpoint validity by requiring every relation endpoint to belong to the canonical node set $V$. Second, we enforce directionality consistency according to the inferred attribute profile $A$: if the figure is predicted as bidirectional or undirected, any valid connection represented by only one directed edge is completed by adding its reciprocal edge. Third, we enforce single-component connectivity by requiring the final graph to contain only one connected component and removing disconnected fragments. The final canonical graph is obtained as
\begin{equation}
G^* = \operatorname{SelfCheck}(V, E_{\text{draft}}, A)
\end{equation}

\begin{figure*}[t]
\centering
\begin{minipage}{0.96\textwidth}
\begin{lstlisting}[
    caption={Prompt used for Stage IV: Topological Consistency Enforcement.},
    label={lst:self_check_prompt},
    basicstyle=\ttfamily\footnotesize,
    breaklines=true,
    breakatwhitespace=false,
    columns=fullflexible,
    keepspaces=true,
    frame=single
]
You are given a canonical node inventory, a draft relationship graph, and inferred global figure attributes.

Task:
Consolidate the draft graph into a globally consistent topology representation without adding new visual evidence.

Inputs:
1. Node inventory and legend summary JSON.
2. Draft relationship JSON.
3. Figure profile JSON.

Rules:
1. Merge redundant or duplicated relation predictions across different focus nodes.
2. Preserve only node names that appear in the provided node inventory.
3. Do not invent new nodes or new relation types.
4. Resolve direction conflicts conservatively using the draft graph and the inferred figure attributes.
5. If the figure profile indicates bidirectional or undirected connectivity, reciprocal relations may be preserved when supported by the draft graph structure.
6. Keep the output as a single consolidated JSON object mapping node names to node data.
7. Preserve relation types using the draft graph and legend terminology when available.

Return JSON only in the following format:
{
  "<node_name>": {
    "relations": [
      {
        "source": "<node_name>",
        "target": "<target_name>",
        "type": "<connection_type_or_empty>"
      }
    ]
  }
}

Do not output markdown, explanations, or any text outside the JSON object.
\end{lstlisting}
\end{minipage}
\end{figure*}

\section{Other Implementation Details}

\subsection{Evaluation Metrics}

We evaluate topology extraction using Node F1 and directed Edge Precision, Recall, and F1. Let $V^{\mathrm{pred}}$ and $V^{\mathrm{gt}}$ denote the predicted and gold node sets, and let $E^{\mathrm{pred}}$ and $E^{\mathrm{gt}}$ denote the predicted and gold directed edge sets.

\paragraph{Node matching.}
Before comparison, node names are normalized by removing formatting artifacts such as hyphens, underscores, and other separator symbols, and matching is performed on the resulting textual form. A predicted node is counted as correct if its normalized node name matches a normalized gold node name. Node precision, recall, and F1 are then computed by set matching over normalized node identities:
\begin{align}
P_{\mathrm{node}} &= \frac{|V^{\mathrm{pred}} \cap V^{\mathrm{gt}}|}{|V^{\mathrm{pred}}|}, \\
R_{\mathrm{node}} &= \frac{|V^{\mathrm{pred}} \cap V^{\mathrm{gt}}|}{|V^{\mathrm{gt}}|}, \\
F1_{\mathrm{node}} &= \frac{2 P_{\mathrm{node}} R_{\mathrm{node}}}{P_{\mathrm{node}} + R_{\mathrm{node}}}.
\end{align}

\paragraph{Directed edge matching.}
A predicted edge is counted as correct only if both endpoint node names and the edge direction exactly match a gold edge. For bidirectional or undirected connections, the gold canonical graph represents them as two directed edges, and edge matching is computed over this directed edge set. Directed edge precision, recall, and F1 are computed as
\begin{align}
P_{\mathrm{edge}} &= \frac{|E^{\mathrm{pred}} \cap E^{\mathrm{gt}}|}{|E^{\mathrm{pred}}|}, \\
R_{\mathrm{edge}} &= \frac{|E^{\mathrm{pred}} \cap E^{\mathrm{gt}}|}{|E^{\mathrm{gt}}|}, \\
F1_{\mathrm{edge}} &= \frac{2 P_{\mathrm{edge}} R_{\mathrm{edge}}}{P_{\mathrm{edge}} + R_{\mathrm{edge}}}.
\end{align}

All reported metrics are macro-averaged over diagrams in each subset.

\subsection{Baseline Implementation Details}
For strong VLM baselines, we used a topology-aware chain-of-thought style prompt that asks the model to carry out a structured reasoning process before producing the final graph JSON. Concretely, the prompt first requires global figure attribute inference, including diagram semantics, link directionality, and link density. It then instructs the model to identify valid labeled nodes, discard non-entity text, preserve original labels, and apply deterministic naming rules, including enclosure-aware disambiguation when necessary. After that, the model is asked to extract visible relationships under explicit directionality and parallel-link constraints, and finally to perform a consistency check to ensure node validity, relation support, and reciprocal-edge correctness when justified by the figure. This prompt design was intended to reduce prompt under-specification and encourage the baselines to reason explicitly about topological structure. All baselines were required to produce outputs under the same JSON schema as our method. For closed-source VLM baselines, we accessed the models through their official APIs. For open-source VLM baselines, we downloaded the released checkpoints from Hugging Face and ran inference locally. For visual reasoning frameworks, we used their official released codebases and followed their original inference pipelines whenever possible. The full prompt is shown in Listing~\ref{lst:baseline_cot_prompt}.

\begin{figure*}[t]
\centering
\begin{minipage}{0.96\textwidth}
\begin{lstlisting}[
    caption={Topology-aware structured prompt used for strong VLM baselines.},
    label={lst:baseline_cot_prompt},
    basicstyle=\ttfamily\footnotesize,
    breaklines=true,
    breakatwhitespace=false,
    columns=fullflexible,
    keepspaces=true,
    frame=single
]
You are given one technical diagram image.

Task:
Analyze the figure and return a structured graph in valid JSON only.

Use the following internal reasoning steps before producing the final answer:

Step 1. Global figure classification
Classify the figure using only explicit evidence:
- diagram_semantics: web | network | structural_diagram
- link_directionality: one_direction | bi_direction
- link_density: multi_link | single_link
Do not guess if evidence is missing.

Step 2. Node identification
- Identify all valid labeled nodes/entities shown in the figure.
- A valid node must have a readable label and participate in at least one visible relation.
- Preserve original labels exactly; do not translate.
- Remove non-entity text such as decorative text, repeated OCR noise, or purely visual annotations.

Naming rules:
- By default, keep the original node label exactly.
- If duplicate node labels exist, append numeric suffixes in reading order (top-to-bottom, then left-to-right).
- If a node is clearly inside a labeled enclosure and disambiguation is necessary, include the enclosure identifier in the node name.
- Use enclosure-aware naming only when the enclosure is explicitly bounded, has a visible label, and contains multiple nodes.
- Do not invent labels for unlabeled enclosures.

Step 3. Relationship extraction
- Extract all visible relationships between valid nodes.
- Follow arrow direction strictly when arrowheads or pointers are visible.
- If the figure is bi_direction and links are undirected or reciprocal, represent both directions.
- If multiple distinct links connect the same node pair, output them separately.
- Include ports or link attributes only when explicitly shown.

Step 4. Consistency check
- Ensure every relation uses valid node names.
- Remove unsupported or hallucinated entities and relations.
- Ensure reciprocal edges are included only when justified by the figure.

Output requirements:
- Return valid JSON only.
- Each node must appear as a top-level key.
- Each node must list its relationships.
- Do not output explanations, markdown, OCR dumps, coordinates, or visual descriptions.

Output format:
{
  "<node_name>": {
    "relations": [
      {
        "source": "<node_name>",
        "target": "<target_node>",
        "type": "<connection_type_or_empty>",
      }
    ]
  }
}
\end{lstlisting}
\end{minipage}
\end{figure*}

\subsection{Models and APIs}

Unless otherwise specified, the main TopoAgent experiments instantiate all VLM modules with GPT-4o through the official API. To evaluate the generality of the framework across different VLM backbones, we additionally instantiate TopoAgent with Qwen2.5-VL-72B and Qwen2.5-VL-7B in the backbone generalization experiments.

\subsection{Inference Settings}

Unless otherwise specified, all baseline experiments use temperature $=0$. Each model is evaluated in a single-pass setting with the topology-aware structured prompt shown in Listing~\ref{lst:baseline_cot_prompt}.

\section{Additional Experiments and Analysis}

\subsection{Generalization Across VLM Backbones}

To examine whether TopoAgent depends on a specific proprietary VLM, we additionally instantiate the framework with Qwen2.5-VL-72B and Qwen2.5-VL-7B. As shown in Table~\ref{tab:backbone_analysis}, TopoAgent improves topology extraction across all evaluated backbones and both benchmark subsets. The gains are particularly pronounced for smaller models on the more challenging Network-style subset, where Qwen2.5-VL-7B improves from 29.6 to 43.7 Edge F1. These results suggest that the benefits of TopoAgent primarily arise from its structured reasoning design rather than reliance on GPT-4o.

\begin{table}[t]
\centering
\small
\resizebox{\linewidth}{!}{
\begin{tabular}{lllcccc}
\toprule
\textbf{Dataset} & \textbf{Backbone} & \textbf{Setting} &
\textbf{Node F1} & \textbf{Edge P} & \textbf{Edge R} & \textbf{Edge F1} \\
\midrule

Web-style & GPT-4o & Single-pass & 93.3 & 37.8 & 30.0 & 32.8 \\
 & GPT-4o & TopoAgent   & 95.6 & 43.9 & 49.9 & 45.8 \\
\cmidrule(lr){2-7}

 & Qwen2.5-VL-72B & Single-pass & 95.5 & 47.5 & 42.3 & 43.8 \\
& Qwen2.5-VL-72B & TopoAgent   & 96.1 & 49.2 & 46.8 & 47.0 \\
\cmidrule(lr){2-7}

 & Qwen2.5-VL-7B & Single-pass & 83.8 & 36.1 & 41.1 & 37.8 \\
& Qwen2.5-VL-7B & TopoAgent   & 89.3 & 39.5 & 44.0 & 39.6 \\

\midrule

Network-style & GPT-4o & Single-pass & 85.2 & 57.8 & 47.1 & 53.5 \\
 & GPT-4o & TopoAgent   & 87.3 & 63.0 & 56.8 & 58.1 \\
\cmidrule(lr){2-7}

& Qwen2.5-VL-72B & Single-pass & 76.0 & 50.7 & 46.9 & 47.5 \\
 & Qwen2.5-VL-72B & TopoAgent   & 85.3 & 59.5 & 54.1 & 55.6 \\
\cmidrule(lr){2-7}

 & Qwen2.5-VL-7B & Single-pass & 63.9 & 37.1 & 26.3 & 29.6 \\
& Qwen2.5-VL-7B & TopoAgent   & 73.2 & 52.2 & 39.1 & 43.7 \\

\bottomrule
\end{tabular}
}
\caption{
Generalization across VLM backbones on the Web-style and Network-style subsets.
We compare single-pass inference with TopoAgent using GPT-4o, Qwen2.5-VL-72B, and Qwen2.5-VL-7B.
All precision, recall, and F1 scores are macro-averaged over diagrams.
}
\label{tab:backbone_analysis}
\end{table}

\subsection{Controlled Baseline with OCR and Region Crops}

To isolate the contribution of structured topology reasoning from additional perceptual evidence, we evaluate a stronger GPT-4o single-pass baseline augmented with the same external OCR results and region crops used by TopoAgent. As shown in Table~\ref{tab:ocr_crop_baseline}, providing these additional inputs substantially improves the single-pass baseline, confirming that localized visual and textual evidence is beneficial. Nevertheless, TopoAgent achieves higher Edge Recall and Edge F1 on both subsets, with a particularly clear advantage on Network-style diagrams. These results suggest that the gains of TopoAgent cannot be explained solely by access to additional OCR or region-level evidence, but also arise from its node-centric relation reasoning and global topology consistency mechanisms.

\begin{table}[t]
\centering
\small

\resizebox{0.88\linewidth}{!}{
\begin{tabular}{lcccc}
\toprule
\multicolumn{5}{c}{\textbf{Web-style}} \\
\midrule
\textbf{Method}
& \textbf{Node F1}
& \textbf{Edge P}
& \textbf{Edge R}
& \textbf{Edge F1} \\
\midrule
GPT-4o single-pass
& 93.3 & 37.8 & 30.0 & 32.8 \\
GPT-4o single-pass + OCR/crops
& 95.6 & 43.1 & 36.9 & 40.5 \\
TopoAgent
& 95.6 & 43.9 & 49.9 & 45.8 \\
\bottomrule
\end{tabular}
}

\vspace{4pt}

\resizebox{0.88\linewidth}{!}{
\begin{tabular}{lcccc}
\toprule
\multicolumn{5}{c}{\textbf{Network-style}} \\
\midrule
\textbf{Method}
& \textbf{Node F1}
& \textbf{Edge P}
& \textbf{Edge R}
& \textbf{Edge F1} \\
\midrule
GPT-4o single-pass
& 85.2 & 57.8 & 47.1 & 53.5 \\
GPT-4o single-pass + OCR/crops
& 86.7 & 61.0 & 48.5 & 54.1 \\
TopoAgent
& 87.3 & 63.0 & 56.8 & 58.1 \\
\bottomrule
\end{tabular}
}

\caption{
Controlled comparison with a GPT-4o single-pass baseline augmented with the same OCR results and region crops used by TopoAgent. All metrics are macro-averaged over diagrams.
}
\label{tab:ocr_crop_baseline}
\end{table}
% \begin{table}[t]
% \centering
% \small
% \resizebox{\linewidth}{!}{
% \begin{tabular}{lcccccccc}
% \toprule
% & \multicolumn{4}{c}{\textbf{Web-style}}
% & \multicolumn{4}{c}{\textbf{Network-style}} \\
% \cmidrule(lr){2-5}
% \cmidrule(lr){6-9}
% \textbf{Method}
% & \textbf{Node F1} & \textbf{Edge P} & \textbf{Edge R} & \textbf{Edge F1}
% & \textbf{Node F1} & \textbf{Edge P} & \textbf{Edge R} & \textbf{Edge F1} \\
% \midrule
% GPT-4o single-pass
% & 93.3 & 37.8 & 30.0 & 32.8
% & 85.2 & 57.8 & 47.1 & 53.5 \\

% GPT-4o single-pass + OCR/crops
% & 95.6 & 43.1 & 36.9 & 40.5
% & 86.7 & 61.0 & 48.5 & 54.1 \\

% TopoAgent
% & 95.6 & 43.9 & 49.9 & 45.8
% & 87.3 & 63.0 & 56.8 & 58.1 \\
% \bottomrule
% \end{tabular}
% }
% \caption{
% Controlled comparison with a GPT-4o single-pass baseline augmented with the same OCR results and region crops used by TopoAgent. All metrics are macro-averaged over diagrams.
% }
% \label{tab:ocr_crop_baseline}
% \end{table}

\subsection{Oracle-Node Edge Extraction}

To disentangle node extraction errors from relation-reasoning errors, we conduct an oracle-node analysis on both benchmark subsets. In this setting, each method is provided with the gold node set and evaluated only on relation prediction, removing node detection and node-name matching errors as confounding factors. As shown in Table~\ref{tab:oracle_node_ablation}, both GPT-4o and TopoAgent benefit substantially from oracle nodes. On Web-style diagrams, Edge F1 improves from 32.8 to 49.5 for GPT-4o and from 45.8 to 57.9 for TopoAgent. On Network-style diagrams, it improves from 53.5 to 64.0 for GPT-4o and from 58.1 to 72.3 for TopoAgent, confirming that node extraction errors propagate into downstream edge prediction.

Importantly, TopoAgent consistently outperforms GPT-4o under both standard and oracle-node settings, suggesting that its gains are not solely due to improved node grounding but also reflect stronger relation reasoning. The substantial improvement of GPT-4o with oracle nodes further indicates that jointly resolving node grounding and relation prediction remains challenging for single-pass VLM inference.

\begin{table}[t]
\centering
\small
\resizebox{\linewidth}{!}{
\begin{tabular}{llccc}
\toprule
\textbf{Dataset} & \textbf{Method} & \textbf{Edge P} & \textbf{Edge R} & \textbf{Edge F1} \\
\midrule

Web-style & GPT-4o single-pass        & 37.8 & 30.0 & 32.8 \\
 & GPT-4o w/ Oracle Nodes    & 48.1 & 53.8 & 49.5 \\
 & TopoAgent                 & 43.9 & 49.9 & 45.8 \\
 & TopoAgent w/ Oracle Nodes & \textbf{56.7} & \textbf{62.5} & \textbf{57.9} \\

\midrule

Network-style & GPT-4o single-pass        & 57.8 & 47.1 & 53.5 \\
 & GPT-4o w/ Oracle Nodes    & 66.8 & 60.7 & 64.0 \\
 & TopoAgent                 & 63.0 & 56.8 & 58.1 \\
 & TopoAgent w/ Oracle Nodes & \textbf{72.1} & \textbf{75.7} & \textbf{72.3} \\

\bottomrule
\end{tabular}
}
\caption{
Oracle-node analysis on the Web-style and Network-style subsets. Gold node sets are provided to each method, and only edge extraction is evaluated. All precision, recall, and F1 scores are macro-averaged over diagrams.
}
\label{tab:oracle_node_ablation}
\end{table}

\subsection{Error-Type Analysis}

To further characterize the remaining topology extraction failures, we categorize errors into four types: direction errors, hallucinated edges, missed edges, and node-name mismatches. Table~\ref{tab:error_breakdown} reports the number and proportion of each error type on the two benchmark subsets.

\begin{table}[t]
\centering
\small
\resizebox{\linewidth}{!}{
\begin{tabular}{lccccc}
\toprule
\textbf{Dataset} & \textbf{Total Errors} & \textbf{Direction} & \textbf{Hallucinated} & \textbf{Missed} & \textbf{Node-name Mismatch} \\
\midrule
Web-style
& 911
& 240 (26.3\%)
& 301 (33.0\%)
& 300 (32.9\%)
& 70 (7.7\%) \\
Network-style
& 1915
& 56 (2.9\%)
& 304 (15.9\%)
& 1072 (56.0\%)
& 483 (25.2\%) \\
\bottomrule
\end{tabular}
}
\caption{
Error-type breakdown of TopoAgent on the Web-style and Network-style subsets. Percentages denote the fraction of total errors within each subset.
}
\label{tab:error_breakdown}
\end{table}

The two subsets exhibit distinct failure patterns. Web-style errors are relatively balanced across direction, hallucination, and missed-edge errors, while Network-style errors are dominated by missed edges (56.0\%) and node-name mismatches (25.2\%). This suggests that Web-style diagrams are more sensitive to local relation interpretation, whereas Network-style diagrams are mainly challenged by topology completeness and node grounding. These trends are consistent with the oracle-node results and further highlight relation recall as a major remaining bottleneck.

\subsection{Inference Efficiency and Cost}

We further analyze the inference overhead of TopoAgent compared with single-pass GPT-4o. Table~\ref{tab:inference_cost} reports the average API calls, inference time, and token usage on 50 randomly sampled benchmark diagrams.

\begin{table}[t]
\centering
\small
\resizebox{\linewidth}{!}{
\begin{tabular}{lccccc}
\toprule
\textbf{Method} & \textbf{Avg. API Calls} & \textbf{Avg. Time} & \textbf{Avg. Prompt Tok.} & \textbf{Avg. Output Tok.} & \textbf{Avg. Total Tok.} \\
\midrule
GPT-4o single-pass      & 1.00  & 14.34s & 1,677.7  & 1,037.8 & 2,715.4  \\
TopoAgent               & 14.04 & 98.05s & 28,350.0 & 3,029.2 & 31,379.4 \\
\bottomrule
\end{tabular}
}
\caption{
Average inference cost on 50 randomly sampled benchmark diagrams. 
Prompt tokens include textual instructions, image-token accounting, and auxiliary contextual information; output tokens correspond to generated JSON results and intermediate responses.
}
\label{tab:inference_cost}
\end{table}

As expected, the multi-stage design introduces additional inference cost compared with single-pass inference, while providing substantial gains in topology extraction performance. This overhead is largely associated with repeated node-centric reasoning and could potentially be reduced through more efficient model routing. For example, future implementations could use smaller open-source VLMs for local node-level reasoning while reserving stronger models for global planning and verification.

\section{Qualitative Examples}
\label{app:qualitative}

\subsection{Successful Cases}
We present two successful examples, one from the Web-style subset and one from the Network-style subset. In both cases, TopoAgent achieves perfect topology extraction, with Node F1 = 1.0 and Edge F1 = 1.0. These examples show that TopoAgent performs particularly well when the diagram structure is relatively simple, the number of nodes is small, node names are unique and clearly identifiable, and the figure contains few irrelevant visual structures.
\begin{figure*}[t]
    \centering
    \includegraphics[width=\textwidth]{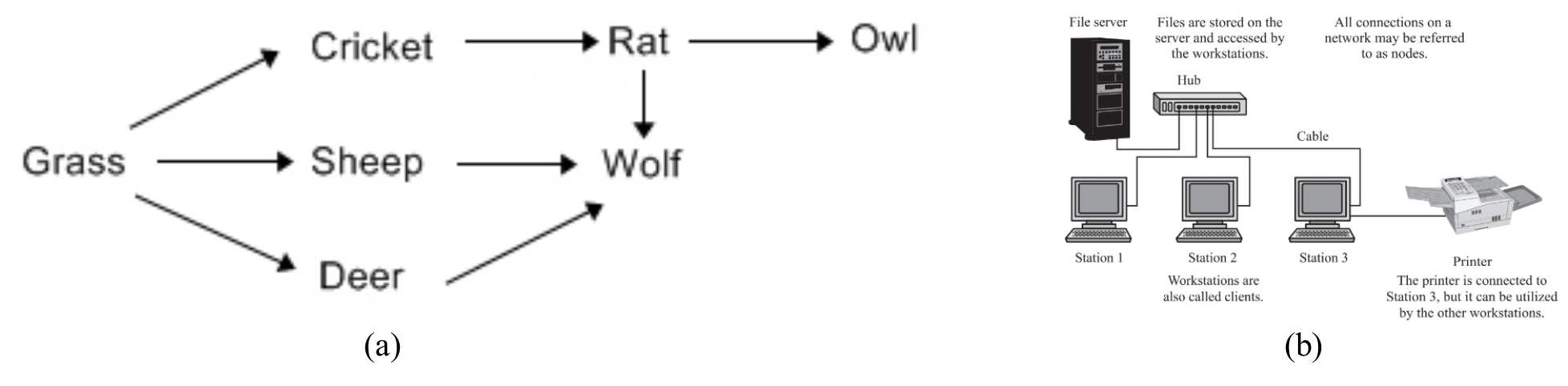}
    \caption{Successful qualitative examples of TopoAgent. The combined figure shows one Web-style example on the left and one Network-style example on the right. In both cases, TopoAgent achieves perfect topology extraction, with Node F1 = 1.0 and Edge F1 = 1.0. These examples show that the framework performs particularly well when the diagram structure is relatively simple, the number of nodes is small, node names are unique and clearly identifiable, and the figure contains few irrelevant visual structures.}
    \label{fig:qualitative_success}
\end{figure*}

\subsection{Challenging Cases}
We present two challenging examples, one from the Web-style subset and one from the Network-style subset. In these cases, TopoAgent achieves Edge F1 scores of 31.3 and 45.0, respectively. These examples illustrate that performance degrades when the diagram contains many nodes, long or crossing arrows, ambiguous directionality, difficult-to-read text, or visually cluttered non-structural elements. Such factors make both entity grounding and directed relation extraction substantially more difficult.

\begin{figure*}[t]
    \centering
    \includegraphics[width=\textwidth]{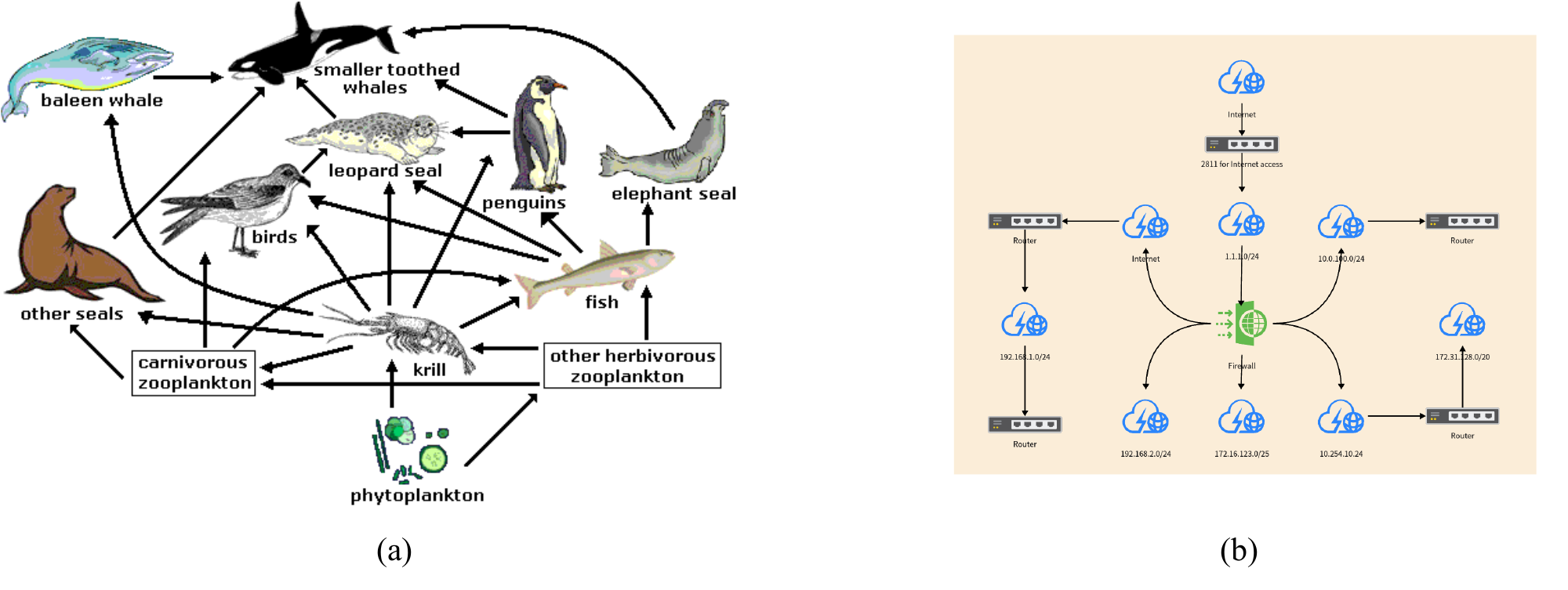}
    \caption{Challenging qualitative examples of TopoAgent. The combined figure shows one Web-style example on the left and one Network-style example on the right. TopoAgent achieves Edge F1 scores of 31.3 and 45.0, respectively. These examples illustrate typical failure factors, including many nodes, long or crossing arrows, ambiguous edge directionality, difficult-to-read text, and visually cluttered structure, all of which increase the difficulty of node grounding and directed edge extraction.}
    \label{fig:qualitative_failure}
\end{figure*}
\section{Use of Large Language Models (LLMs)}

Large language models (LLMs) and other AI-based tools were used during the preparation of this manuscript to assist with writing refinement, language polishing, and code editing. All AI-assisted outputs were carefully reviewed, verified, and revised by the authors. The authors take full responsibility for the final content of the manuscript.
% \section{Prompts and Structured Output Templates}
% \label{app:prompts}

% \subsection{Region Summarization Prompt}
% \subsection{Figure Classification Prompt}
% \subsection{Node Inventory Prompt}
% \subsection{Relationship Prompt}
% \subsection{Self-Check Prompt}

% \section{Experimental Setup}
% \label{app:exp}

% \subsection{Models and APIs}
% \subsection{Inference Settings}
% \subsection{Baseline Configurations}
% \subsection{Evaluation Protocol}

% \section{Qualitative Examples}
% \label{app:qualitative}

% \subsection{Successful Cases}
% \subsection{Failure Cases}
% \subsection{Comparison with Representative Baselines}

% \section{Ethical Considerations and Data Licensing Details}
% \label{app:ethics}

% \subsection{Data Sources and Redistribution}
% \subsection{Potential Annotation Bias}
% \subsection{Limitations of Deployment}

\appendix
\end{document}